\pdfoutput=1
\documentclass[10pt,twocolumn,letterpaper]{article}
\usepackage{cvpr} % PAPER TYPE
\usepackage{graphicx}
\usepackage{multirow}
\usepackage{amsmath}
\usepackage{amssymb}
\usepackage{booktabs}
\usepackage[ruled,vlined]{algorithm2e}
\usepackage{verbatim}
\usepackage[pagebackref=false,breaklinks=true,colorlinks,bookmarks=false]{hyperref}
\usepackage[capitalize]{cleveref}
\crefname{section}{Sec.}{Secs.}
\Crefname{section}{Section}{Sections}
\Crefname{table}{Table}{Tables}
\crefname{table}{Tab.}{Tabs.}
\def\cvprPaperID{6237}
\def\confName{CVPR}
\def\confYear{2022}

\begin{document}
%%%%%%%%% TITLE
\title{Continual Learning for Visual Search with Backward Consistent Feature Embedding}

\author{Timmy S. T. Wan\textsuperscript{1} \qquad Jun-Cheng Chen\textsuperscript{2} \qquad Tzer-Yi Wu\textsuperscript{3} \qquad Chu-Song Chen\textsuperscript{1,\thanks{\;indicates corresponding author.}}\\
National Taiwan University\textsuperscript{1} \qquad Academia Sinica\textsuperscript{2} \qquad ucfunnel Co. Ltd.\textsuperscript{3}\\
{\tt\small \{r08944004,chusong\}@csie.ntu.edu.tw, pullpull@citi.sinica.edu.tw, kenny.wu@ucfunnel.com}
}
\maketitle

%%%%%%%%% ABSTRACT
\begin{abstract}
In visual search, the gallery set could be incrementally growing and added to the database in practice. However, existing methods rely on the model trained on the entire dataset, ignoring the continual updating of the model. Besides, as the model updates, the new model must re-extract features for the entire gallery set to maintain compatible feature space, imposing a high computational cost for a large gallery set. 
To address the issues of long-term visual search, we introduce a continual learning (CL) approach that can handle the incrementally growing gallery set with backward embedding consistency.
We enforce the losses of inter-session data coherence, neighbor-session model coherence, and intra-session discrimination to conduct a continual learner.
In addition to the disjoint setup, our CL solution also tackles the situation of increasingly adding new classes for the blurry boundary without assuming all categories known in the beginning and during model update. 
To our knowledge, this is the first CL method both tackling the issue of backward-consistent feature embedding and allowing novel classes to occur in the new sessions.
Extensive experiments on various benchmarks show the efficacy of our approach under a wide range of setups\footnote{Code: https://github.com/ivclab/CVS}.
\end{abstract}

%%%%% Figure 1
\begin{figure}[t]
 \begin{center}
  \includegraphics[width=0.85\columnwidth]{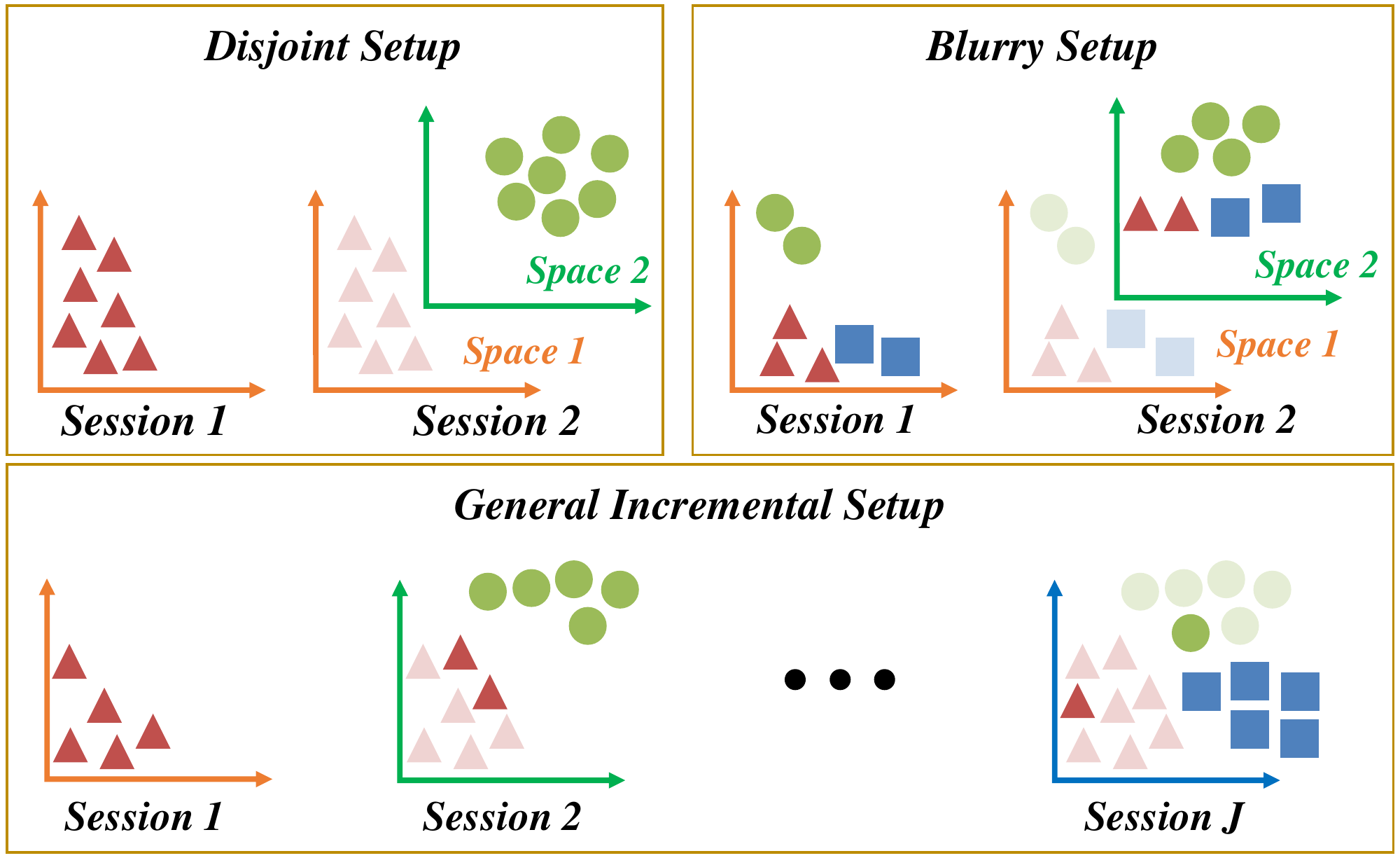}
 \end{center}
  \caption{Illustration of the proposed approach with the \emph{general incremental setup}. 
  Our solution allows the new gallery set of seen and unseen classes to be freely and incrementally added to the database with respect to the widely adopted \emph{disjoint} and the recent \emph{blurry setups}. In addition, it also considers backward compatible embedding for a session sequence. This avoids the gallery embeddings of old and new sessions from being separated in incompatible feature spaces. 
  Thus, our approach is more practical for real retrieval applications. Semi-transparent icons represent the data points from the previous sessions collected so far.}
  \label{fig:setups}
\end{figure}

%%%%%%%%% INTRODUCTION
\section{Introduction}
\label{sec:intro}

Continual learning (CL) aims to learn new tasks while keeping the functions learned from the old sessions.
The technology has been rapidly evolving;
nevertheless, the active research area in CL focuses on image classification but ignores the demand for image retrieval (aka~visual search).
For obtaining powerful feature representations in image retrieval, most works~\cite{Musgrave2020AML,pmlr-v119-roth20a,Wang_2019_CVPR,Zhai2019ClassificationIA} still require a model to be trained on an entire dataset simultaneously instead of in an incremental manner.
However, a practical visual search system should be capable of continually learning from new materials while consolidating the old knowledge to cope with the data accumulated with time.

As data grows, despite updating the model by simply fine-tuning, many observations~\cite{ChenWei2020Oteo,Zhao2021ContinualRL} reveal that catastrophic forgetting happens.
A series of strategies have been developed in CL to address the problem~\cite{ChenWei2020Oteo,Rebuffi_2017_CVPR,Li2018LearningWF}.
The methods can make a single deep model capable of updating itself successively while avoiding disappointing overall performance.
However, there are still several ongoing issues.

First, many works in CL emphasize the disjoint setup where the data from the old class will not show during training in a new task (or session). 
The task boundary arising over classes restricts the usage of CL since many retrieval systems need to collect extra data of the seen labels for improving their models in new sessions. 
Although recent studies (\eg \cite{Bang_2021_CVPR,Aljundi2019GradientBS}) allow the class overlapping among the tasks, the blurry setup in these works assumes that all the class labels in the future sessions are pre-given in advance; only the instance ratios in the classes vary with the session (Fig.~\ref{fig:setups}). 
This kind of data scenario is impractical for most visual-search applications (e.g., in an e-commerce system, the new product arrives over season). 
Moving toward a more general setup is desirable to fulfill real-world scenarios.

Second, a model updated from new data will deploy online in retrieval.
An essential step is to re-extract the feature embedding from previous gallery images to maintain a consistent feature space on the pairwise distance measurement.  
For visual search on large-scale data, feature re-extraction is computationally intensive. 
Thus, an ideal design for continual visual search is that an updated model only extracts features for incoming gallery data while keeping the previously generated feature representations unchanged.
However, it leads to the further difficulty of pairwise similarity measurement in uneven feature spaces.
Motivated by this, we argue that current CL studies lack consideration for feature compatibility between the ongoing and previous data.
Hence, designing a CL algorithm with backward consistent feature embedding is demanded.

To address the above issues, we introduce a novel CL approach, namely, CVS (\textbf{C}ontinual-learner for \textbf{V}isual \textbf{S}earch), for a generally incremental setup of visual search.
CVS can learn effective feature representations with backward consistency.
For learning new knowledge, our learner obtains discriminating features for the current task. 
We introduce a cross-task gallery embedding consistency constraint that keeps the currently learned features compatible with the representatives from the obsolete gallery features.
For consolidating old knowledge while keeping feature consistency, we develop a metric-based knowledge distillation to draw together the embedding from the different feature spaces. 
By coordinating the components, CVS can achieve continual visual search with backward-compatible feature embedding effectively. 
Also, we conduct extensive experiments under various incremental data distributions, especially for the general-incremental setup, to validate the effectiveness of CVS.
The main characteristics include:

\noindent \textbf{General Incremental Setup}: We introduce a new CL scenario to simulate real-world visual search application systems.
It includes the previous disjoint and blurry setups as special cases and tackles the general setting that the classes in a coming session can be either seen or unseen before.

\noindent \textbf{Backward Consistent Feature Space Learning}: Our learner can learn discriminative features for unseen classes. 
It can also maintain the distance metric learned effective for the seen classes in both new and old sessions, where the old-session features can be kept unchanged without the need to be re-extracted every time in a visual search system.

\begin{figure}
  \includegraphics[width=1.02\linewidth]{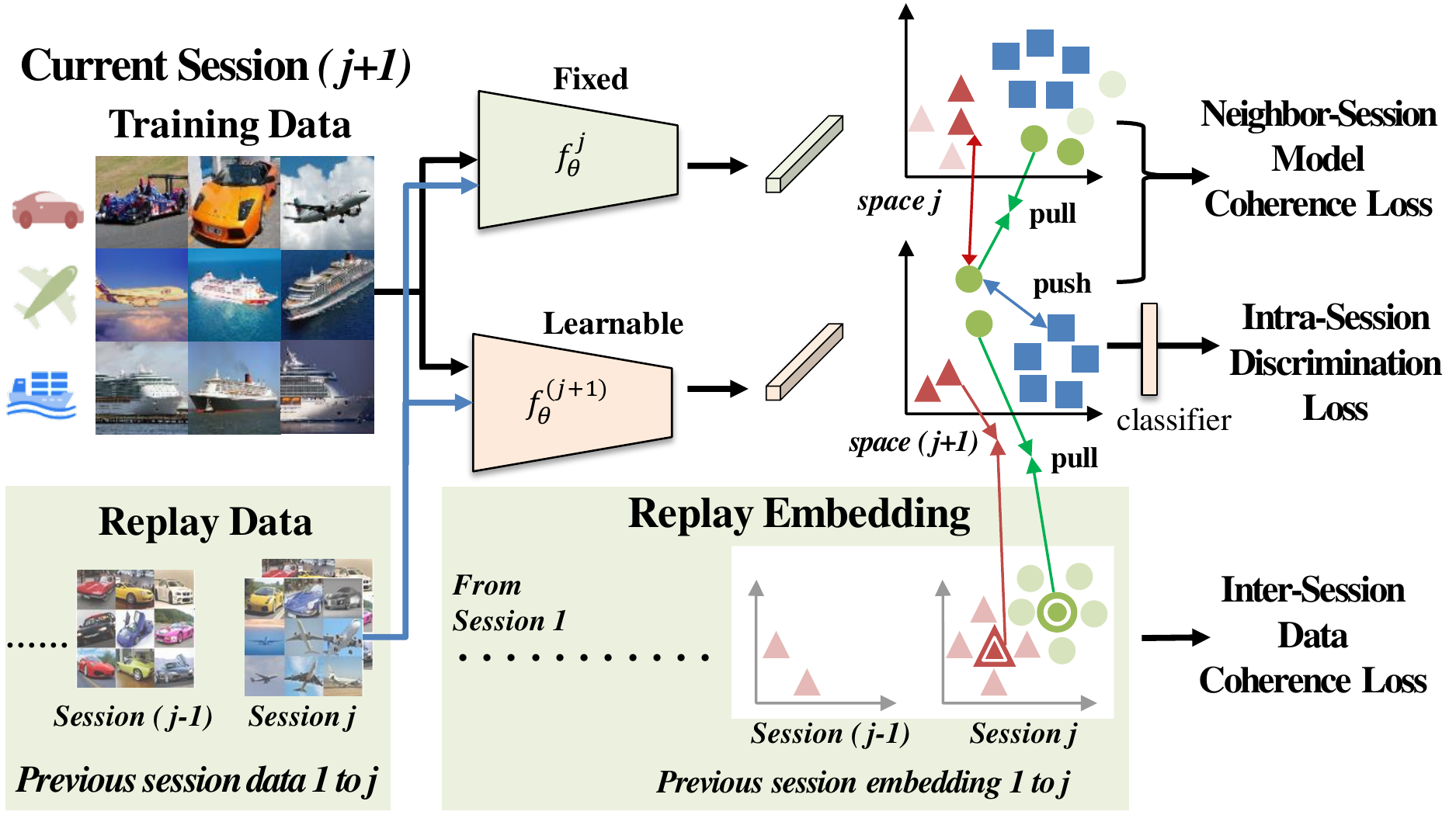}
  \caption{Overview of the proposed approach for CL in General-Incremental setup with backward embedding consistency for a long-term learning. We enforce the three losses for a continual learner: intra-session discrimination loss to learn discriminative representation with mainly the current-session data, neighbor-session model coherence loss to regulate the current model with the previous-session model for backward compatible embedding, and inter-session data coherence with mainly the current-session data and replayed embedding of all previous sessions for long-term embedding consistency.
  Semi-transparent icons represent the data points from the previous sessions collected so far.
  Note that we omit the replayed data to simplify the illustration.}
  \label{fig:continual_senario}
\end{figure}

Extensive experiments on multiple datasets under incremental data distributions show that our method achieves state-of-the-art results.
Fig.~\ref{fig:continual_senario} shows our system diagram.

%%%%%%%%%%%%%%%%%%%% FIGURE OVERVIEW OF CVS

%%%%%%%%%%%%%%%% RELATED WORK
\section{Related Work}
\label{sec:related_work}

We briefly review the recent progress of image retrieval in Section \ref{section2.1}.
We then summarize the CL and describe its challenge on similarity-based visual search in Section~\ref{section_review_CL}.

\subsection{Image Retrieval}\label{section2.1}

Image retrieval ranks the gallery images in order given the query image.
Previous studies rely on either descriptors by local feature aggregation~\cite{Sivic2003VideoGA,5540009,5540039} or low-level visual clues~\cite{1039125,993558,531803} and then perform a nearest neighbor search.
The modern way utilizes embeddings from the neural network (\ie, neural mapping) instead because the learnable descriptors show superior accuracy~\cite{Schroff2015FaceNetAU,Sohn2016ImprovedDM,Song2016DeepML,MovshovitzAttias2017NoFD,Chen_2017_CVPR,Zhai2019ClassificationIA} and compact storage~\cite{7301269,Cao_2017_ICCV,Yuan_2018_ECCV,Yuan_2020_CVPR}.
To derive the neural mapping, metric learning optimizes the model in a pairwise manner or pointwise manner.
Pairwise methods derive the discriminating space where positive pairs become closer and negative pairs repel each other.
A fair benchmark~\cite{pmlr-v119-roth20a} shows the state-of-the-art results of the pairwise method. 
However, an issue is that the complexity of the sample mining grows in magnitude when the number of samples in a group is beyond two, (\eg, triplet loss~\cite{Schroff2015FaceNetAU} and quadruplet~\cite{Chen_2017_CVPR}).
Even though many studies~\cite{Hermans2017InDO,Wu_2017_ICCV,Wang_2019_CVPR,Wang_2020_CVPR,Xuan_2020_WACV} delve into the mining issue, sampling the informative pairs is intrinsically hard without clear indicators.
On the other hand, pointwise methods~\cite{Wan_2014_ACMMM,MovshovitzAttias2017NoFD,Zhai2019ClassificationIA,Qian2019SoftTripleLD,Teh2020ProxyNCARA,Kim_2020_CVPR} treat the optimization as data points against representative samples per class.
Instead of meticulous mining, the representative samples could be learnable proxies~\cite{MovshovitzAttias2017NoFD,Qian2019SoftTripleLD,Teh2020ProxyNCARA,Kim_2020_CVPR} or randomly sampled points~\cite{Wan_2014_ACMMM,Zhai2019ClassificationIA}, bringing faster convergence for training without pairwise comparisons.
For example, NSoftmax~\cite{Zhai2019ClassificationIA} follows the classification training paradigm and applies L2 normalization to the embedding layer output for image retrieval.
Such a minor modification yields competitive results compared with pairwise methods in the benchmark~\cite{pmlr-v119-roth20a}.
Our method is easily integrated with the strategies mentioned above. For simple and elegant purposes, we optimize with NSoftmax.

\noindent\textbf{Backward Consistency of Feature Embedding}:
Despite the proliferation of different methods in image retrieval, most work ignores the demand for backward consistency, \ie, making previously frozen features comparable with the newly extracted ones in the gallery set.
Feature consistent learning contributes to this goal in two ways.
The first line aims to reduce the effort of feature re-extraction.
R³AN~\cite{Chen_2019_CVPR} projects the old features into the new feature space by one-side transform; CMC~\cite{WangCYCL_2020_BMVC} bridges the multiple feature spaces via a lightweight transformation module.
However, they still need re-extraction, which becomes impractical for large galleries and long sessions.
The second line maintains the backward consistency without any feature re-extraction.
BCT~\cite{Shen_2020_CVPR} constrains the current feature space by simultaneously enabling gradient flow from both the old and the new classifier. 
However, it requires the entire previous data seen so far for finetuning and hence is incapable of handling the long session learning scenario.
Moreover, the gallery set is assumed fixed in the experimental settings of BCT without addressing the critical issue that the gallery set could be increasingly enlarged in practice.
Unlike BCT, our method imposes constraints on both the inter-session feature embeddings and the neighbor-session model; hence it gains a great improvement in backward consistency. 
Besides, we examine an at most 10-session scenario (in contrast to only 3 sessions at most in BCT's experiments) for a reality check on the backward feature compatibility.

\subsection{Continual Learning}\label{section_review_CL}

CL aims for a single learner that can sequentially update knowledge without forgetting the previously learned information.
The existing works~\cite{van2019three,Hsu18_EvalCL} can be categorized into task-incremental and class-incremental CL, where the former assumes the task index presented at the inference time while the later assumes a task-agnostic scenario during inference.
Most works assume the categories in the successive tasks (\ie, sessions) to be disjoint to each other.

To avoid overfitting to the current session, regularization-based methods~\cite{Kirkpatrick2017OvercomingCF,Zenke2017ContinualLT,Chaudhry_2018_ECCV,schwarz2018progress} impose constraints on change of the critical neural weights, distillation-based methods~\cite{Rannen_2017_ICCV,Li2018LearningWF,Dhar_2019_CVPR,Zhang2020ClassincrementalLV,douillard2020podnet} transfer the knowledge from the previous learner to the current one, replay-based methods~\cite{Rebuffi_2017_CVPR,Brahma2018SubsetRB,hayes2019memory,Liu_2020_CVPR,Bang_2021_CVPR} revisit a small amount of old data to prevent forgetting, and isolation-based methods~\cite{Mallya_2018_CVPR,Mallya_2018_ECCV,pmlr-v80-serra18a,hung2019compacting} assign the subnetwork capacity per task.
Though an automatic task selector~\cite{Abati_2020_CVPR} can help extend an isolation-based method to class-incremental, it still requires the disjoint assumption of the classes among tasks.
To break the disjoint-class limitation in CL, recent studies propose the blurry setup (\ie, class-overlapping~\cite{Aljundi2019GradientBS,prabhu2020greedy,Bang_2021_CVPR}).
However, it requires the presentation of all predefined classes across subsequent sessions, and is thus still impractical for real-world visual search applications.

Besides, most works focus on the classification problem, ignoring the demand for image retrieval. 
MMD~\cite{ChenWei2020Oteo} studies retrieval-based CL in a distillation manner.
By narrowing down the mean difference between two distributions in reproducing kernel space, the current model distills the knowledge of the old one.
However, the method ignores the backward-compatible needs for retrieval; hence the embeddings have to be re-extracted for the gallery data as the task grows in CL.
It is thus impractical for a realistic system.

%%%%%%%%%%%%%%%%%%%%%% METHOD
\section{Continual Learner for Visual Search (CVS)}
\label{sec:method}

Consider a CL retrieval problem with $J$ sessions.
In session $j\in\{1\cdots J\}$, a neural-network model is learned and let $f_j$ be the mapping from the input image to the feature embedding of the model.
During CL training, the backbone feature extractor $f_{j+1}$ is initialized from $f_{j}$, the neural mapping obtained immediately before the current session.
To fulfill the requirement in our General-Incremental setup, the gallery set is incrementally extensible with the newly added embedding. Specifically, let $G_{j}$ be the 
images newly added in time session $j$ and $\mathbf{g}_j$ be the feature embedding obtained for $G_{j}$ (via $f_j$). 
The embedding will be added to the gallery set up to the $j$-th session, $\mathbf{g}_{1:j}=\bigcup_{i=1}^j\mathbf{g}_i$.  
The feature embedding cumulatively saved (\ie,~$\mathbf{g}_{1:j}$) then serve as the gallery set for retrieval for future sessions $l(>j)$. In addition, to learn $f_{j+1}$ capable of handling the new-session data containing possibly both existing and novel classes and also maintaining the performance on all previous-session data, we conduct the following loss terms in our CVS approach, namely, inter-session term of data coherence ($\text{\ensuremath{\mathbf{L}}}_{\{1:j\};j+1}^{\text{\ensuremath{\mathbf{d}}}}$),
neighbor-session term for model coherence ($\text{\ensuremath{\mathbf{L}}}_{j;j+1}^{\text{\ensuremath{\mathbf{m}}}}$),
and intra-session term for discrimination ($\text{\ensuremath{\mathbf{L}}}_{j+1}^{\text{\ensuremath{\mathbf{c}}}}$).

%%%%%%%%%%%%%%%%%%% 3.1 Inter-session data coherence
\subsection{Inter-session data coherence }

Each training sample of the previous-session data has been already 
provided with an embedding by one of the old neural mappings, $f_{1},\cdots,f_{j}$.
To fulfill the condition of not re-extracting the embedding for updating the gallery set every time, they are fixed and kept invariant once having been built upon a session $k\in\{1\cdots j\}$. 
That is, after establishing the embedding, they can be used for a series of future tasks in CL.
Besides hoping to un-forget the classification power of known labels, we further need to un-forget the capability of retrieval based on the embedding already built.

\noindent\textbf{Replayed Embedding Together with Data}: 
Replayed data are widely adopted to avert forgetting in CL.
To resolve storage efficiency and avoid training on all data, a small portion of the data from sessions $1$ to $j$ can be stored and replayed for a joint training with the current data in session $j+1$.
In our work, the old data embeddings have been extracted as $\mathbf{g}_i=f_i\{G_i\} |_{i=1}^j$.
Besides replaying the data, we replay the embedding to facilitate CL, which is more efficient with learning.
We call the technique \emph{replayed embedding}, where a small portion of features sampled from the embedding space are replayed for training.

The deep model updated in the current session $j+1$ should be confined to the stored embedding of seen classes.
Therefore, the goal of our learner is to train $f_{j+1}$ to extract features from $G_{j+1}$ ($\textbf{g}_{j+1}=f_{j+1}\{G_{j+1}\}$) while keeping the existing obsolete features $\mathbf{g}_{1:j}$ unchanged.
However, the embeddings are established from different neural mappings varying with sessions.
The sessions have separated distributions and could be diverse for a long sequence of sessions.
As different feature spaces are not necessarily comparable, to avoid an over-fitting to individual sessions, we aggregate the embedding across the previous sessions by taking the expectations as follows.

\begin{equation}
\mathcal{E}_c=\frac{1}{j}\sum_{i=1}^{j}\mathrm{\xi}(\mathbf{g}_{ic}), c\in \mathcal{C}(i),
\end{equation}
where $\mathcal{C}(i)$ is the set of class indices appearing in session $i$, $\textbf{g}_{ic}$ is the set of embedding extracted from the data of class $c$ in $G_i$, and $\mathrm{\xi}(\cdot)$ denotes the expectation operator. 
The loss enforcing inter-session data coherence is defined as follows:
\begin{equation}\label{eqn:loss_center_1}
\mathbf{L}_{\{1:j\};j+1}^{\mathbf{d_I}} = \sum_{c\in\Pi_{j+1}} \sum_{x_i\in c} \left \| f_{j+1}(x_{i})-\mathcal{E}_c \right \| _{2}^{2},
\end{equation} 
where $\Pi_{j+1}=\bigcup_{i=1}^j\mathcal{C}(i) \cap \mathcal{C}(j+1)$ is the indices intersection between the classes in the current and all previous sessions; $x_i$ denotes the data in the current session ($j+1$).

However, the current-session classes may not contain all the previous classes in our General Incremental setup.
\Eg, for the special case of Disjoint setup, there is no old class sample in the incoming sessions at all.
Hence, besides employing the replayed embedding $\mathcal{E}_c$ (in $\Pi_{j+1}$), 
we use the replayed data for the old classes $\Gamma_{j+1}=\bigcup_{i=1}^j\mathcal{C}(i)$ too.
Without loss of generality, we use the exemplar mining technique iCaRL~\cite{Rebuffi_2017_CVPR} to conduct a small portion of data for replay, which searches random neighbors around the mean per class.
The replayed embedding and data in $\Gamma_{j+1}$ are co-used as follows:
\begin{equation}\label{eqn:loss_center_2}
\mathbf{L}_{\{1:j\};j+1}^{\mathbf{d_O}} = \sum_{c\in\Gamma_{j+1}} \sum_{\tilde{x}_i\in c} \left \| f_{j+1}(\tilde{x}_{i})-\mathcal{E}_c \right \| _{2}^{2},
\end{equation}
where $\tilde{x}_i$ denotes the replayed data.
Then, the inter-session data coherence loss is a combination of Eqs.~\ref{eqn:loss_center_1} and ~\ref{eqn:loss_center_2}, which helps maintain the backward feature consistency:
\begin{equation}\label{eqn:loss_center}
\mathbf{L^d}_{\{1:j\};j+1} = (\mathbf{L}_{\{1:j\};j+1}^{\mathbf{d_I}}+\mathbf{L}_{\{1:j\};j+1}^{\mathbf{d_O}})/n,
\end{equation}
where $n$ is the mini-batch size.

The loss combines both the replayed embedding and data to maintain the coherence of the feature spaces of sessions $1, \cdots, (j+1)$.
Even the aggregated embeddings provide strong constraints, as the freedom of neural networks is often large in the capacity, the learner can still easily adapt the model from $f_j$ to $f_{j+1}$.
In our experience, a nice usage of this precedent constraint can help build an effective CL learner because the fixed embeddings can act as attractors to regularize the deep model training.
Our approach only needs $|\Gamma_{j+1}|$ replayed embeddings, and we set a ratio of around $5\%$ samples of the total data in sessions $1:J$ as a fixed budget shared by all sessions up to now in the replayed data.
Since the loss establishes the connection between session $(j+1)$ to all previous ones $(1:j)$, unlike BCT~\cite{Shen_2020_CVPR} employing only the inter relationship between the current session $(j+1)$ and that immediately before the session $(j)$, our approach can effectively enhance the backward feature consistency for long-term sessions.

Besides, our replayed-embedding solution coincides with the idea in the cross-batch learning~\cite{YangACMMM2016,Wang_2020_CVPR}, where previously learned embeddings in the past mini-batches guide the training in the current one.
The approaches have been demonstrated effective to boost retrieval performance.
Our approach is analogously a cross-\textit{session} learning method.
In the experiments, we show the efficacy of our cross-session CL solution and make an in-depth analysis on the results.

%%%%%%%%%%%%% 3.2 Neighbor-session model coherence
\subsection{Neighbor-session model coherence}

In the above, we utilize the replayed materials of seen classes in all previous sessions and the current session data for training. 
This section develops the loss using mainly the current-session data while co-using the current model ($f_{j+1}$) and last-time model ($f_j$) in training.
It enforces the neighbor-session model coherence.

The loss is designed using the distillation principle, where the mapping $f_l$ (learned and fixed) serves as the teacher model to guide the training of the student $f_{l+1}$.
Unlike previous studies of distillation techniques that are mainly for classification~\cite{Hinton_2015_NIPSW}, we conduct a distillation-based loss to regulate the metric learning of $f_{j+1}$ from the metric space of $f_j$ based on the triplet loss to better fit the nature of CL on retrieval. 
In a triple of $(x_a,x_p,x_n)$ for an anchor, a positive sample, and a negative sample respectively are selected from mainly the current-session data. The purpose of training is to shrink the distance between $x_a$ and $x_p$ while enlarging that between $x_a$ and $x_n$.

To distill knowledge from the embedding space already built from the previous session $j$, we generate the embeddings of positive and negative samples using $f_j(\cdot)$ and obtain $f_j(x_p)$ and $f_j(x_n)$, respectively.
Then, we use the teacher-generated embedding to guide the training of the student anchor $f_{j+1}(x_a)$. 
The current session data are fed into the current and previous models to constrain the embedding distributions produced by $f_{j+1}$ and $f_j$, respectively. The loss conducted can thus enforce the model coherence between neighbor sessions.

Besides, we co-use $x_a$ and $x_p$ via setting $x_p=x_a$.
Hence, we conduct a 2-sample-3-embedding triplet-loss strategy where the triplet embeddings are $f_{j+1}(x_a)$, $f_j(x_a)$, and $f_j(x_n)$.
This is because it can facilitate drawing the embedding distributions of the student and the teacher based on the same $x_a$ for comparison.
In the meantime, it also saves the computation of choosing positive samples and enforces efficient sampling.
Recent studies have shown that mining easy positive samples (\ie, similar positives) benefits to metric learning~\cite{Xuan_2020_WACV}.
Our approach directly forms the positive sample's embedding from the anchor, which eliminates the positive mining effort in~\cite{Xuan_2020_WACV} during pair sampling.
For the negative, we follow the hardest negative mining principle~\cite{Hermans2017InDO} to pick the embedding.
Denote $d_j^{j+1}(x,y)=\| f_{j+1}(x)-f_j(y)\|_{2}^{2}$.
The neighbor-session-model-coherence loss is written as:
\begin{equation}\label{eqn:loss_distillation}
\mathbf{L}_{j;j+1}^{\mathbf{m}} = \frac{1}{n} \sum_{x_a}\left [ d_j^{j+1}(x_a,x_a)  - d_j^{j+1}(x_a,x_n) + m  \right ]_{+},
\end{equation}
where $m$ is the margin set as $0.1$ by default.
Besides, $f_{j+1}(x)$ and $f_j(x)$ are $l_2$-normalized to alleviate the issue of forgetting according to previous studies~\cite{Hou_2019_CVPR,DBLP:conf/icann/LuoZXWRY18}.
This loss helps restrict the behavior of the updated model coherent to the previous one. 

%%%%%%%%%%%%%%%%%%%%%% 3.3 Intra-session discrimination
\subsection{Intra-session discrimination}

Recent studies show that classification is a strong baseline for learning effective feature embedding (if an appropriate $l_2$-norm normalization layer is added)~\cite{Zhai2019ClassificationIA}. 
Without loss of generality, we employ the method to establish the retrieval capability of $f_{j+1}$ using mainly the current-session data.
The loss for intra-session discrimination is
\begin{equation}\label{eqn:loss_classification}
\mathbf{L}_{j+1}^{\mathbf{c}} = \frac{1}{n} \sum_{i=1}^{n}-{log(\frac{exp(w_{y_{i}}^{T}f_{j+1}(x_{i})/T)}{\sum_{k}exp(w_{k}^{T}f_{j+1}(x_{i})/T) } )},
\end{equation}
where $(x_i,y_i)|_{i=1}^n$ are the data and labels, $w$ denote the $l_2$-normalized weight of the classification layer, $f_{j+1}(x_{i})$ is the $l_2$-normalized embedding extracted for $x_i$, and $T$ is the temperature term set as $0.05$ by default.

%%%%%%%%%%%%%%%%%%%%%% Total Loss
\noindent\textbf{Total Loss.}
The final learning objective is as follows:
\begin{equation}\label{eqn:loss_combination}
    \min_{f_{j+1}}{\mathbf{L}} = \mathbf{L}_{j+1}^{\mathbf{c}} + \alpha \mathbf{L}_{j;j+1}^{\mathbf{m}} + \beta \mathbf{L^d}_{\{1:j\};j+1}.
\end{equation}
By default, we set $\alpha$ to $10$ and $\beta$ to $1$ empirically in our experiments, and cosine distance is used for retrieval in this work.
Only the first term $\mathbf{L}_{1}^{\mathbf{c}}$ is used in session $1$.
Then, all three terms are used in sessions $2$ to $J$.
Fig.~\ref{fig:continual_senario} gives an illustration of our approach.

In sum, to fulfill the General Incremental setup, the first term $\mathbf{L}_{j+1}^{\mathbf{c}}$ provides a fundamental retrieval ability of the current session ($j+1$).
The second term $\mathbf{L}_{j;j+1}^{\mathbf{m}}$ makes the neural mapping $f_{j+1}$ close to $f_j$ based on both the novel- and seen-label data in the current session.
It enforces the prediction to mimic the behavior of the previous feature extractor.
The third term $\mathbf{L^d}_{\{1:j\};j+1}$ enforces the backward feature consistency from the replayed embedding and data of all sessions ($1$ to $j$) and a joint training with the current-session data based only on the labels seen before (\ie, $\bigcup_{i=1}^j\mathcal{C}(i)$).
It helps bias the feature space to align with the one obsolete gallery features lie in.
By joining the three losses, our CVS approach can handle the General Incremental setup and its special cases (Blurry, Disjoint) well.

%%%%%%%%%%%%%%%%%%%%%%%%% EXPERIMENT
\section{Experiments}
\label{sec:experiments}

We conduct extensive experiments across various datasets under different data distributions. 
Five datasets are used, including two \emph{coarse-grained} datatests, CIFAR100 and Tiny ImageNet, and three \emph{fine-grained} datasets, Stanford Dog, iNaturalist 2017, and Product-10K.
We use CIFAR100 for fundamental study and then Tiny Imagenet on a longer sequence of sessions.

The datasets are summarized as follows.
\textbf{CIFAR100} \cite{Krizhevsky09learningmultiple} has 100 categories; each owns 500 $32\times32$ images for training and 100 images for testing.
\textbf{Tiny ImageNet}~\cite{Le2015TinyIV} has 100,000 training and 10,000 testing images (of size $64\times64$ sampled from the 200 classes from ImageNet.
Due to its compact size and various categories, we conduct a long sequence learning on it.
\textbf{Stanford Dog}~\cite{KhoslaYaoJayadevaprakashFeiFei_FGVC2011} contains 120 dog breed-level categories picked from ImageNet, including 8,580 images for testing and 12,000 images for training. 
For the training split, each class has 100 images.
\textbf{iNaturalist 2017}~\cite{Horn_2018_CVPR} is a large-scale long-tailed image retrieval dataset with 5,089 species-level categories.
We sample 527 images per class for the 200 classes; each contains at least 527 samples to avoid very small classes and unbalanced data before partition.
We call it \textbf{iNat-M} in this work.
\textbf{Product-10K}~\cite{bai2020products} is an ultrafine-grained long-tailed dataset covering the top-9691 frequently bought product images from a real e-commerce system; Each class's images contain diverse appearances of the specific product by collecting offline customer-taken and online in-shop photos. 
We remove the category with fewer than 20 images in the training set.
We then construct the train-test split based on the original training set because the official testing data do no offer labels. 
Finally, there are 2,743 classes.
We call it \textbf{Product-M} in this work.
Fine-grained images are more challenging as only subtle differences exist.

%%%%%%%%%%%%%%%%%%% 4.1 Implementation details
\subsection{Implementation Details}
In all of the experiments, the embedding dimension sets as 128 by default. 
We use ResNet-18~\cite{He_2016_CVPR} for coarse-grained datasets, and ResNet-50~\cite{He_2016_CVPR} for fine-grained datasets.
The hyperparameter details are depicted in the supplementary material.

\begin{figure}[t]
    \centering
            \begin{subfigure}[]{1.0\linewidth}
                    \centering
                    \includegraphics[width=\textwidth]{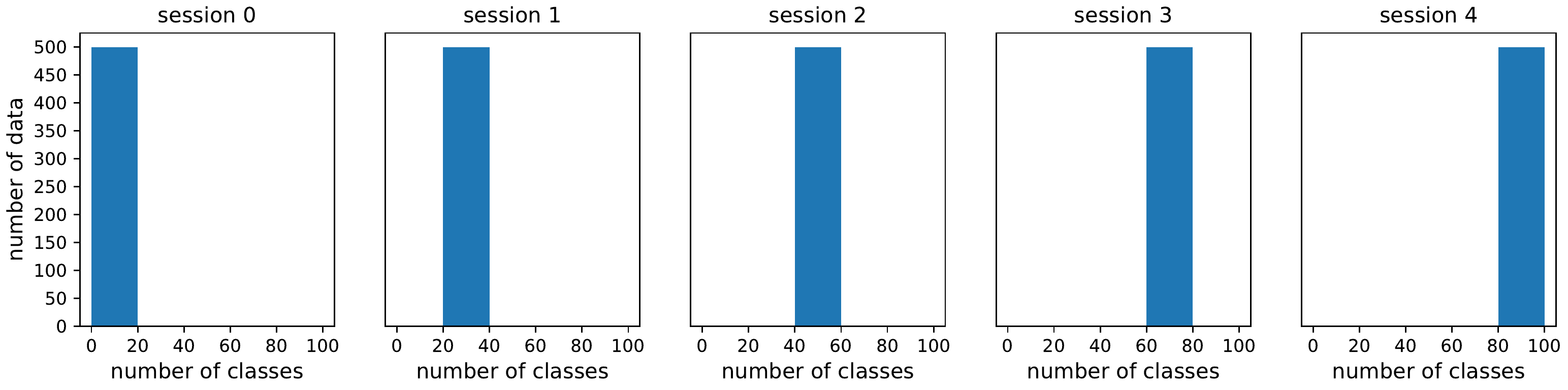}
                    \caption{Disjoint Setup}\label{fig:cifar100distributiondisjoint}
            \end{subfigure}
            
            \begin{subfigure}[]{1.0\linewidth}
                    \centering
                    \includegraphics[width=\textwidth]{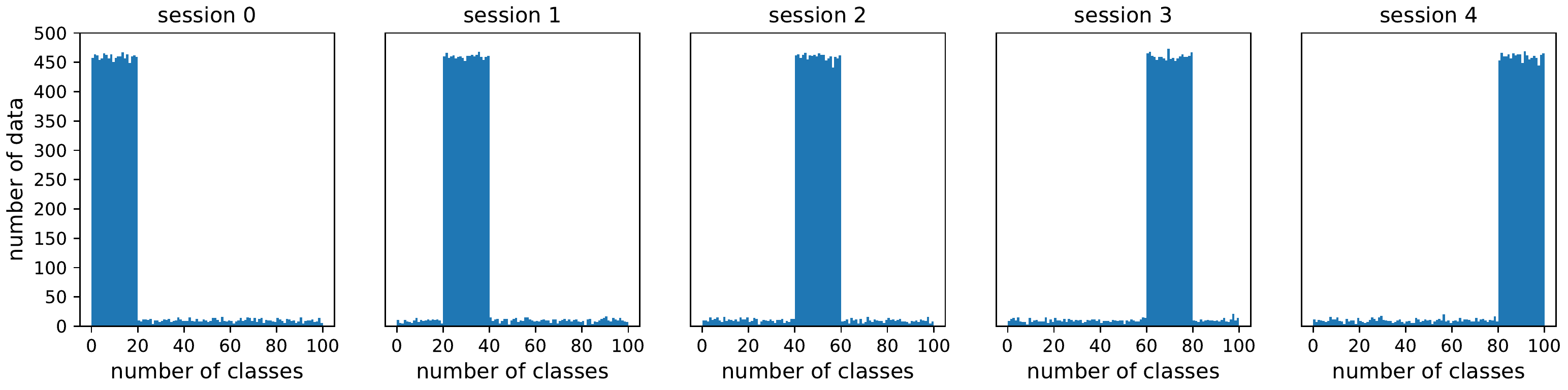}
                    \caption{Blurry Setup}\label{fig:cifar100distributionblurry}
            \end{subfigure}
            
            \begin{subfigure}[]{1.0\linewidth}
                    \centering
                    \includegraphics[width=\textwidth]{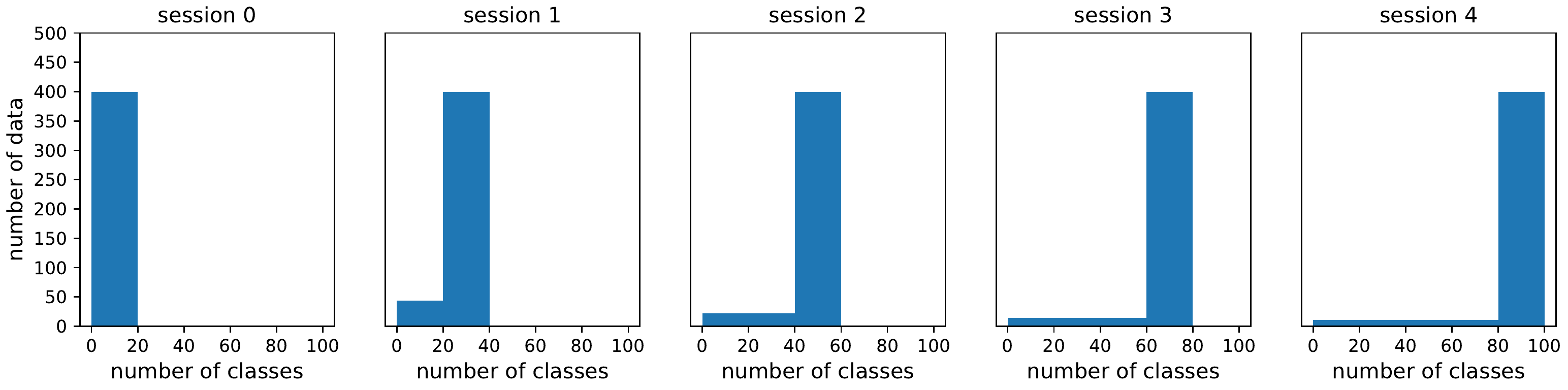}
                    \caption{General Setup}\label{fig:cifar100distributiongeneral}
            \end{subfigure}
    \caption{Class distributions of different setups on CIFAR100. }
    \label{fig:cifar100distribution}
\end{figure}

\noindent\textbf{Disjoint setup} assumes that previously seen classes are unavailable when learning new session data. As shown in Fig.~\ref{fig:cifar100distributiondisjoint}, we partition the CIFAR100 into five sessions with each containing 20 classes. For Tiny ImageNet, the dataset is divided into ten sessions with 20 classes per session.

\noindent\textbf{Blurry setup}~\cite{Bang_2021_CVPR} provides all 
categories in the beginning; each session shows a subset of samples from all of them. 
The data distribution is controlled via a percentage of the data from the major and minor classes per session.
In our experiments, CIFAR100 is divided into five sessions; each contains 20 major classes and 80 minor classes, with 90\% samples from the major and 10\% from the minor, as shown in~\ref{fig:cifar100distributionblurry}.
For Tiny ImageNet, we simulate a 10-session CL; each has 70\% from the major and 30\% from the minor.

\noindent\textbf{General-incremental setup}:
The new-class and novel old-class samples may co-exist. 
The model is learned from the $S$ classes at initial in a total of $L$ sessions.
For the later sessions, we add $C$ classes and assume that $M\%$ of the data are from the old categories and $(1-M)\%$ are from the new, denoted as $(S,C,M,L)$. 
We set $(20,20,10,5)$ for CIFAR100 (Fig.~\ref{fig:cifar100distributiongeneral}). 
For the rest, we apply $(20,20,30,10)$ to Tiny ImageNet, $(60,20,30,4)$ to Stanford Dog, $(100,25,30,5)$ to iNat-M, and (1343,700,40,3) to Product-M.

Replayed data are used for all loss terms. 
Following~\cite{Bang_2021_CVPR}, we set the memory buffer size for replay as 2,000 for CIFAR100.
Note that this is a budget-limited buffer (for the replay-based methods) co-used by all sessions.
We assume a memory budget of 600 samples for Dog due to the small scale, 4,000 for Tiny ImgNet, 4,000 for iNat-M, and 3,000 for the ultrafine-grained Product-M, respectively. 
It is pretty challenging for Product-M as it has 2,743 classes; for each session, only $1\sim2$ images will be stored for replay.

%%%%%%%%%%%%%%%%% FIGURE ABOUT Distribution on CIFAR100

%%%%%%%%%%%%%%%%%%%%% 4.2 Evaluation
\subsection{Evaluation}

We follow the train-validation-test protocol. 
The model achieving the best recall@1 in the validation phase is picked for fair comparisons. 
Then, we report the final performance by using the original testing set as a query set.
For collecting the validation query set for each class in disjoint and general-incremental setups, we sample 5\% of current training session data for iNat-M and Tiny ImageNet, and 10\% for CIFAR100 and Stanford Dog. 
We sample 2 images randomly from each class for Product-M due to scarce training data. 
Note that such a validation query set is accumulated as the session grows in the disjoint and general setup. 
For the blurry setup, we keep a fixed portion of data from the entire training set as a validation query set (5\% for Tiny ImageNet and 10\% for CIFAR100) class-wisely since the model can see all predefined categories in the beginning.
The embedding extracted for the current-session training data is added to the gallery set after finishing the session as mentioned before; thus the new-classes images first observed in one season are allowed to be searched in the future seasons.
Such an extensible gallery set matches the real-world scenarios.

We report the \textbf{recall@$k$}~\cite{Jgou2011ProductQF} as it is the most popular metric in fine-grained retrieval \cite{Song2016DeepML,pmlr-v119-roth20a}.
For the testing query, we evaluate all the classes for the blurry setup, and the classes seen so far for both the disjoint and general setup, respectively. 
Finally, we average the scores over sessions, denoted as \textbf{AR@K}.
To compare with existing approaches, we re-implement the LWF~\cite{Li2018LearningWF}, and MMD~\cite{ChenWei2020Oteo} and re-run EWC~\cite{Kirkpatrick2017OvercomingCF}, RWalk~\cite{Chaudhry_2018_ECCV}, and Rainbow~\cite{Bang_2021_CVPR} from the official Github of~\cite{Bang_2021_CVPR}. 
Among them, all (except MMD) are for classification since CL is rare to be stuided on retrieval yet, 
and we replace all the plain softmax loss with normalized softmax loss~\cite{Zhai2019ClassificationIA} for them for a fair comparison.
We also re-implement BCT~\cite{Shen_2020_CVPR} that is a backward feature compatible approach for retrieval.
We apply the same exemplar mining as ours for BCT in disjoint setup as the Naïve BCT is not a CL solution and requires old classes to pass through.

Besides, we provide direct fine-tuning as the \textbf{lower bound}; in contrast, we offer joint training of all gallery images seen so far allowing re-extraction as the \textbf{upper bound}.

%%%%%%%%%%%%%%%%%%%%%%%% TABLE 1
\begin{table}
\begin{center}
\resizebox{\columnwidth}{!}{
\begin{tabular}{|l|r|r|r|r|r|r|r|r|r|}
\hline
\multirow{2}{*}{} & \multicolumn{3}{c|}{Disjoint}                                                  & \multicolumn{3}{c|}{Blurry}                                                    & \multicolumn{3}{c|}{General}                                                   \\ \cline{2-10} 
                  & \multicolumn{1}{l|}{AR@1} & \multicolumn{1}{l|}{AR@2} & \multicolumn{1}{l|}{AR@4} & \multicolumn{1}{l|}{AR@1} & \multicolumn{1}{l|}{AR@2} & \multicolumn{1}{l|}{AR@4} & \multicolumn{1}{l|}{AR@1} & \multicolumn{1}{l|}{AR@2} & \multicolumn{1}{l|}{AR@4} \\ \hline
Joint Train       & 83.39                    & 85.69                    & 87.64                    & 55.1                     & 57.74                    & 60.33                    & 81.97                    & 84.6                     & 86.82                    \\ \hline
Finetune          & 42.03                    & 43.23                    & 44.74                    & 32.54                    & 36.8                     & 41.11                    & 60.79                    & 64.73                    & 68.14                    \\ \hline \hline
BCT               & 60.24                    & 64.38                    & 68.37                    & 31.71                    & 35.69                    & 39.61                    & 58.31                    & 62.2                     & 65.66                    \\ \hline
LWF               & 49.33                    & 53.51                    & 58.01                    & 36.79                    & 41.63                    & 46.52                    & 65.53                    & 70.91                    & 75.31                    \\ \hline
MMD               & 49.62                    & 53.58                    & 57.87                    & 32.83                    & 36.73                    & 40.64                    & 65.51                    & 70.22                    & 74.33                    \\ \hline
EWC               & 43.6                     & 45.65                    & 48.12                    & 34.75                    & 40.85                    & 46.71                    & 60.89                    & 64.86                    & 68.2                     \\ \hline
RWalk             & 64.11                    & 67.33                    & 70.56                    & 41.78                    & 45.67                    & 49.4                     & 69.9                     & 73.37                    & 76.39                    \\ \hline
Rainbow           & 62.27                    & 65.09                    & 67.43                    & 41.47                    & 44.99                    & 48.27                    & 68.4                     & 71.56                    & 74.2                     \\ \hline \hline
\textbf{Ours (CVS)}              & \textbf{71.47}                    & \textbf{74.8}                     & \textbf{77.51}                    & \textbf{47.47}                    & \textbf{49.86}                    & \textbf{52.17}                    & \textbf{73.95}                    & \textbf{76.73}                    & \textbf{78.84}                    \\ \hline
\end{tabular}
}
\small{(a)}
\resizebox{\columnwidth}{!}{
\begin{tabular}{|l|r|r|r|r|r|r|r|r|r|}
\hline
\multirow{2}{*}{} & \multicolumn{3}{c|}{Disjoint}                                                  & \multicolumn{3}{c|}{Blurry}                                                    & \multicolumn{3}{c|}{General}                                                   \\ \cline{2-10} 
                  & \multicolumn{1}{l|}{AR@1} & \multicolumn{1}{l|}{AR@2} & \multicolumn{1}{l|}{AR@4} & \multicolumn{1}{l|}{AR@1} & \multicolumn{1}{l|}{AR@2} & \multicolumn{1}{l|}{AR@4} & \multicolumn{1}{l|}{AR@1} & \multicolumn{1}{l|}{AR@2} & \multicolumn{1}{l|}{AR@4} \\ \hline
Joint Train       & 54.42                    & 58.62                    & 62.31                    & 35.46                    & 39.29                    & 43.19                    & 47.45                    & 51.94                    & 55.99                    \\ \hline
Finetune          & 19.73                    & 20.92                    & 21.81                    & 18.63                    & 23.17                    & 28.11                    & 31.11                    & 35.87                    & 40.8                     \\ \hline \hline
BCT               & 34.27                    & 37.92                    & 41.22                    & 18.54                    & 23                    & 27.66                    & 30.17                    & 34.64                    & 39.29                    \\ \hline
LWF               & 24.24                    & 27.5                     & 30.67                    & 20.06                    & 25.11                    & 30.52                    & 32.22                    & 38.23                    & 44.56                    \\ \hline
MMD               & 25.83                    & 29.22                    & 32.75                    & 18.51                    & 22.65                    & 27.23                    & 33.35                    & 38.21                    & 43.06                    \\ \hline
EWC               & 19.95                    & 21.89                    & 23.88                    & 17.33                    & 21.42                    & 26.07                    & 27.86                    & 32.67                    & 37.78                    \\ \hline
RWalk             & 32.85                    & 37.09                    & 41.09                    & 22.4                     & 26.11                    & 29.94                    & 33.83                    & 38.43                    & 42.97                    \\ \hline
Rainbow           & 32.6                     & 35.51                    & 38.35                    & 22.64                    & 26.47                    & 30.4                     & 37.53                    & 41.64                    & 45.44                    \\ \hline \hline
\textbf{Ours (CVS)}              & \textbf{38.87}                    & \textbf{42.03}                    & \textbf{45.08}                    & \textbf{26.62}                    & \textbf{29.19}                    & \textbf{31.86}                    & \textbf{38.78}                    & \textbf{42.38}                    & \textbf{45.89}                    \\ \hline
\end{tabular}
}
\small{(b)}
\resizebox{\columnwidth}{!}{
\begin{tabular}{|l|r|r|r|r|r|r|r|r|r|}
\hline
\multirow{2}{*}{} & \multicolumn{3}{c|}{Dog}                                                       & \multicolumn{3}{c|}{iNat-M}                                                    & \multicolumn{3}{c|}{Product-M}                                                 \\ \cline{2-10} 
                  & \multicolumn{1}{l|}{AR@1} & \multicolumn{1}{l|}{AR@2} & \multicolumn{1}{l|}{AR@4} & \multicolumn{1}{l|}{AR@1} & \multicolumn{1}{l|}{AR@2} & \multicolumn{1}{l|}{AR@4} & \multicolumn{1}{l|}{AR@1} & \multicolumn{1}{l|}{AR@2} & \multicolumn{1}{l|}{AR@4} \\ \hline
Joint Train       & 86.98                    & 91.08                    & 93.99                    & 75.85                    & 79.97                    & 83.65                    & 79.36                    & 83.83                    & 87.73                    \\ \hline
Finetune          & 82.7                     & 88.32                    & 92.23                    & 67.75                    & 72.68                    & 77                       & 70.99                    & 76.26                    & 81.16                    \\ \hline \hline
BCT               & 81.73                    & 87.49                    & 91.55                    & 67.34                    & 72.07                    & 75.99                    & 70.48                    & 75.67                    & 80.47                    \\ \hline
LWF               & 83.25                    & 89.29                    & 93.04                    & 68.51                    & 73.71                    & 78.12                    & 72.95                    & 78.38                    & 83.1                     \\ \hline
MMD               & 83.2                     & 88.98                    & 92.71                    & 68.58                    & 73.48                    & 77.79                    & 72.89                    & 78.21                    & 82.96                    \\ \hline
EWC               & 81.64                    & 88.7                     & 92.82                    & 66.3                     & 71.4                     & 75.82                    & 66.01                    & 72.53                    & 78.24                    \\ \hline
RWalk             & 82.41                    & 88.31                    & 92.43                    & 68.77                    & 73.82                    & 78.16                    & 68.71                    & 74.64                    & 80.13                    \\ \hline
Rainbow           & 82.78                    & 89.04                    & \textbf{93.23}                    & 68.68                    & 73.22                    & 77.21                    & 69.39                    & 75.19                    & 80.34                    \\ \hline \hline
\textbf{Ours (CVS)}              & \textbf{84.71}                    & \textbf{89.4}                     & 92.61                    & \textbf{72.57}                    & \textbf{76.39}                    & \textbf{79.87}                    & \textbf{75.47}                    & \textbf{80.36}                    & \textbf{84.68}                    \\ \hline
\end{tabular}
}
\small{(c)}
\vspace{-15pt}
\end{center}
\caption{Results on (a) CIFAR100 and (b) Tiny ImageNet, and (c) Fine-grained datasets, where \emph{Joint Train} and \emph{Finetune} specify the theoretically upper and lower bounds, respectively.}
\label{tbl:averagerecall}
\end{table}

%%%%%%%%%%%%%%%%%%%%%%%%%%%% FIGURE ABOUT RESULTS ON CIFAR100
\begin{figure*}[t]
    \centering
            \begin{subfigure}[t]{0.3\textwidth}
                    \includegraphics[width=\textwidth]{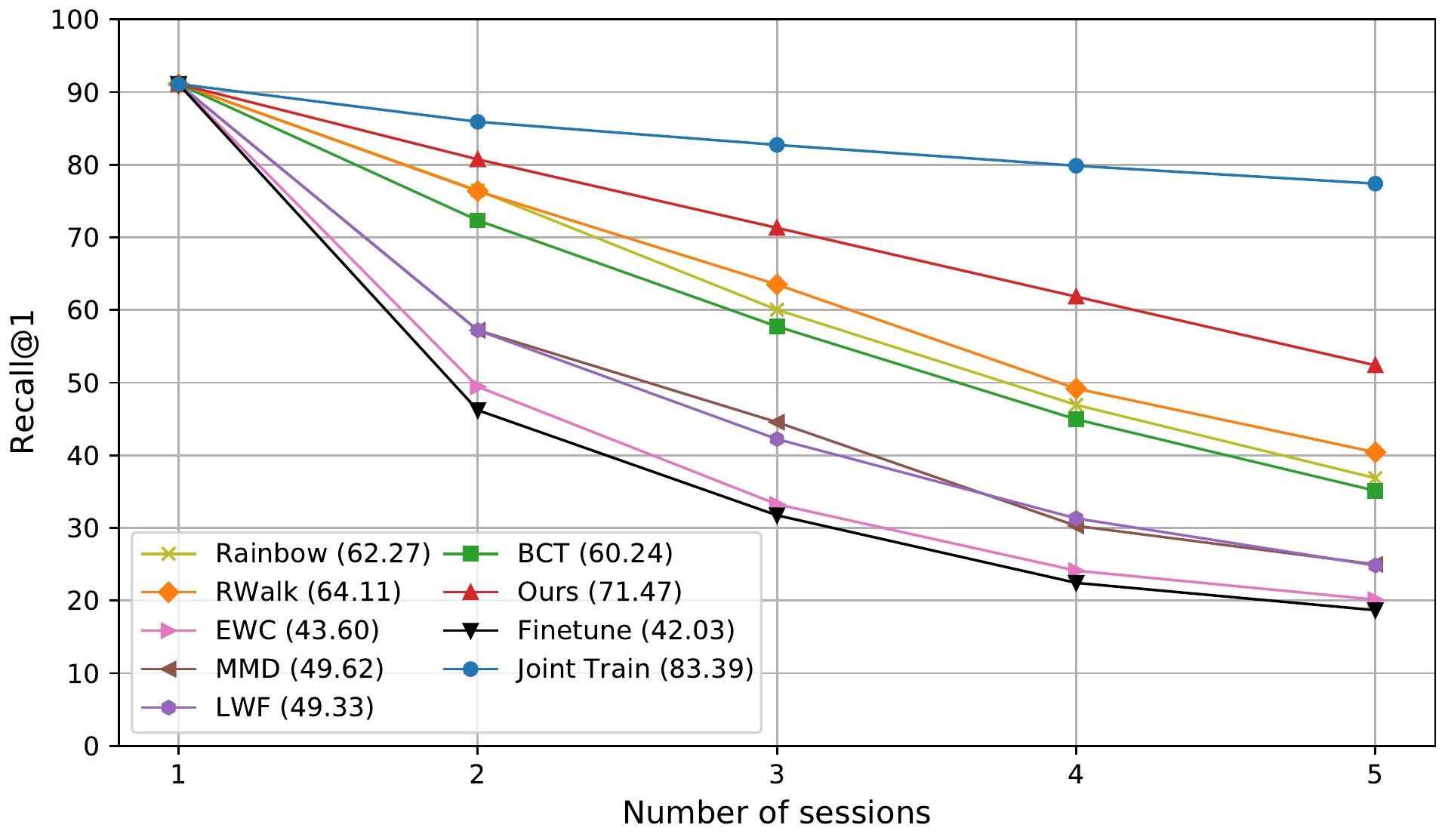}
                    \caption{Disjoint Setup}\label{fig:cifar100figuresdisjoint}
            \end{subfigure}
            \begin{subfigure}[t]{0.3\textwidth}
                    \includegraphics[width=\textwidth]{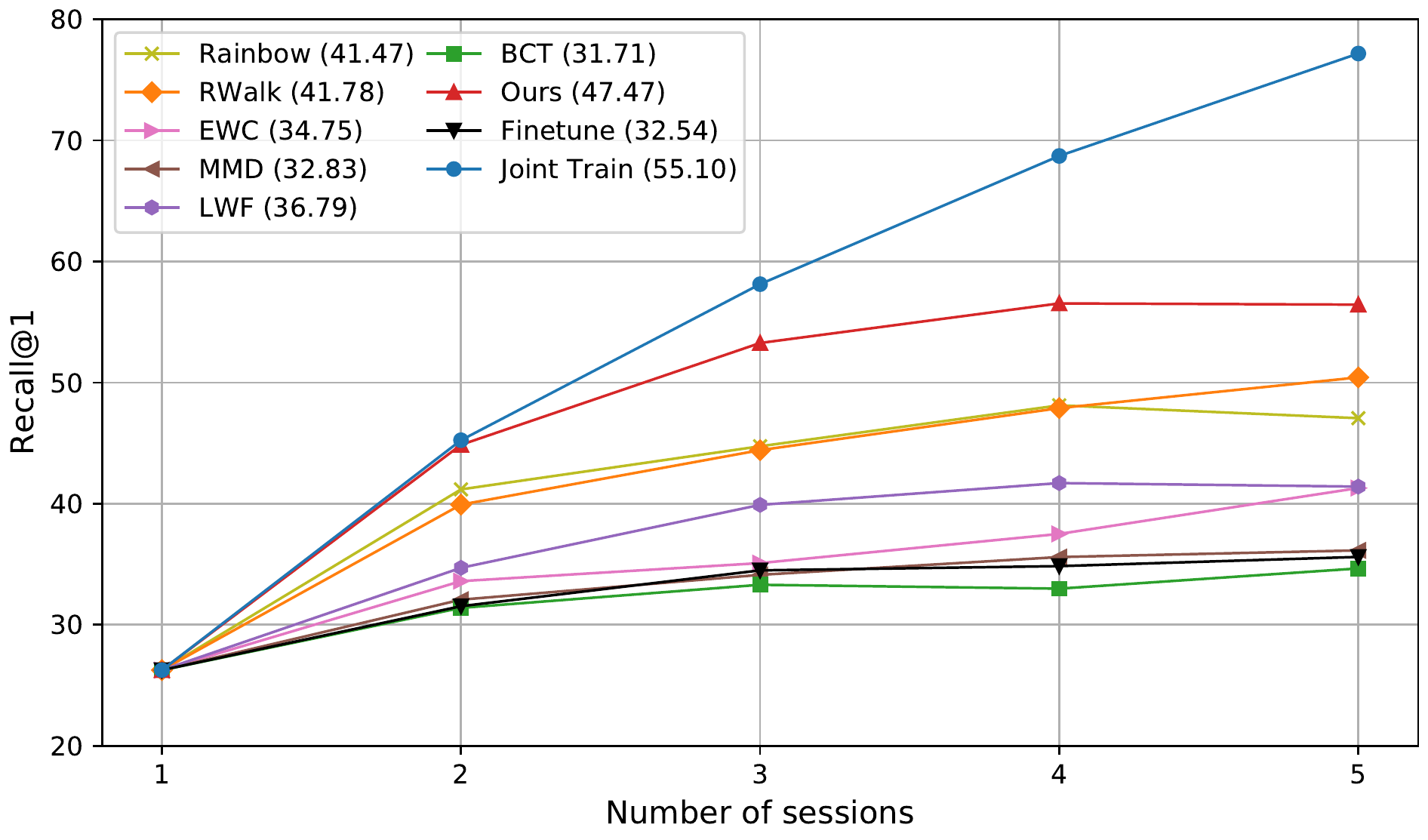}
                    \caption{Blurry Setup}\label{fig:cifar100figuresblurry}
            \end{subfigure}
            \begin{subfigure}[t]{0.3\textwidth}
                    \includegraphics[width=\textwidth]{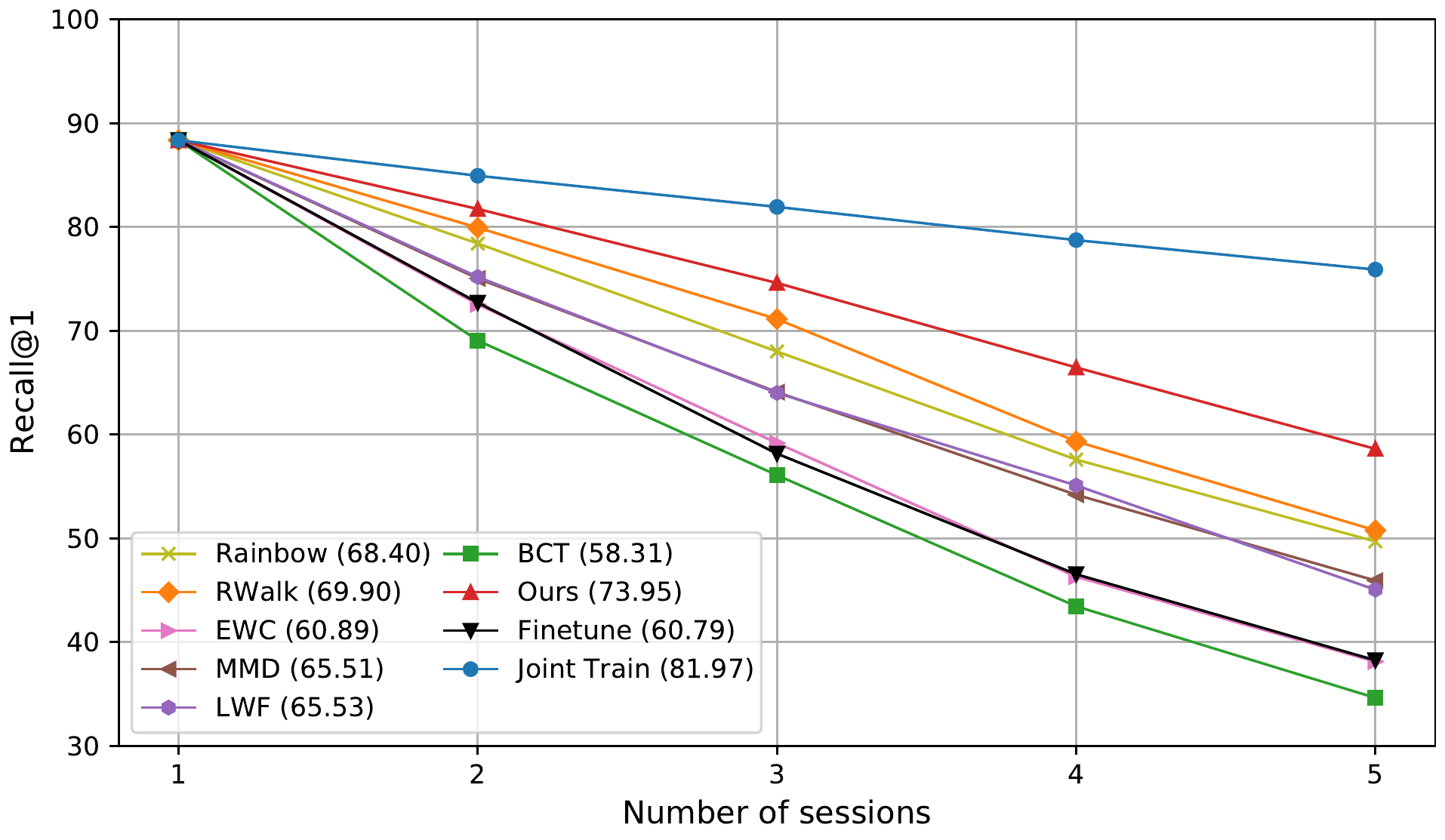}
                    \caption{General Setup}\label{fig:cifar100figuresgeneral}
            \end{subfigure}
    \vspace{-6pt}
    \caption{Recall@1 on CIFAR100 for the three setups, where Average Recall@1 across sessions is reported in parentheses. }
    \label{fig:cifar100figures}
\end{figure*}

%%%%%%%%%%%%%%%%%%%%%%%%%%% FIGURE ABOUT RESULTS ON TINY IMAGENET
\begin{figure*}[t]
    \centering
            \begin{subfigure}[t]{0.3\textwidth}
                    \includegraphics[width=\textwidth]{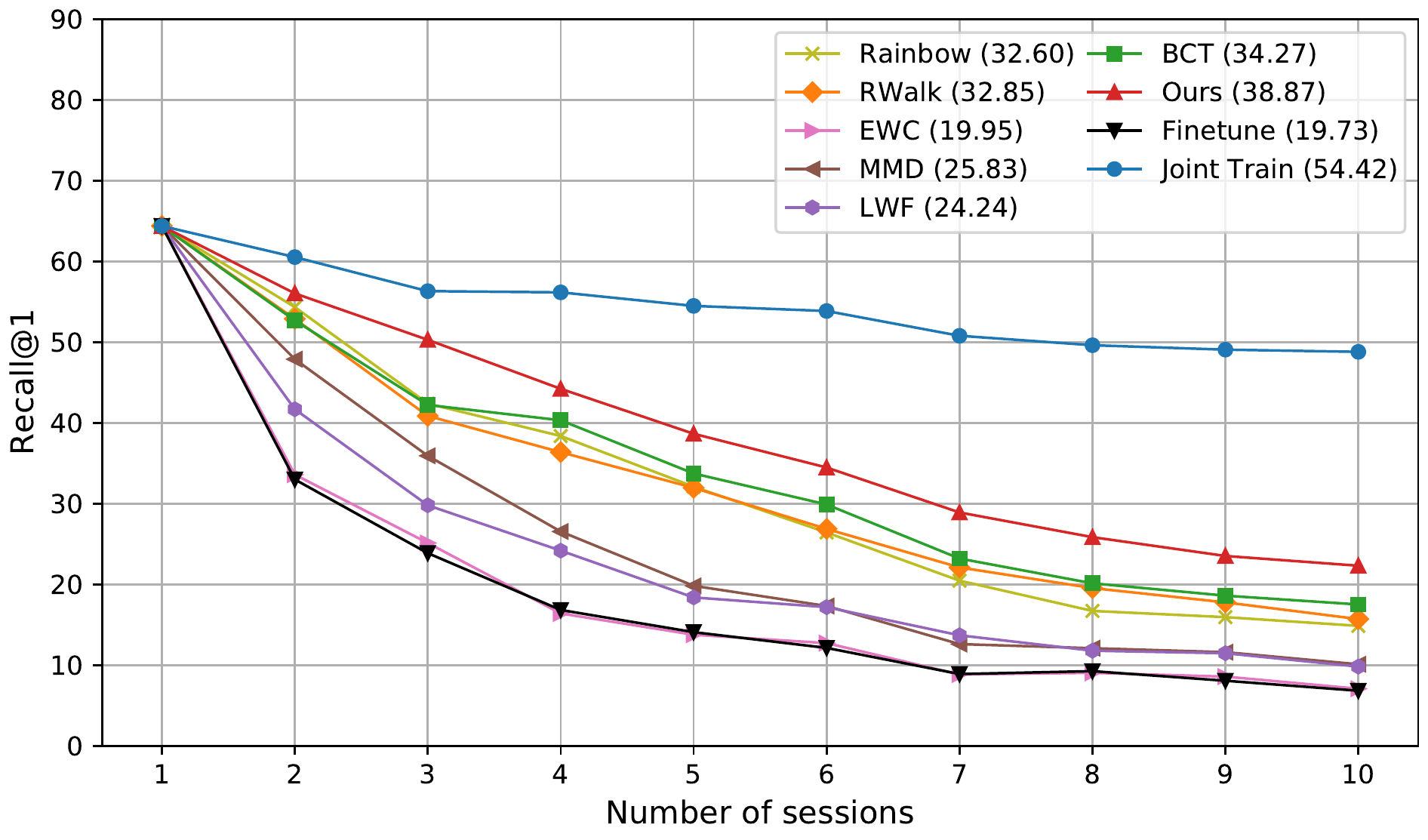}
                    \caption{Disjoint Setup}
            \end{subfigure}
            \begin{subfigure}[t]{0.3\textwidth}
                    \includegraphics[width=\textwidth]{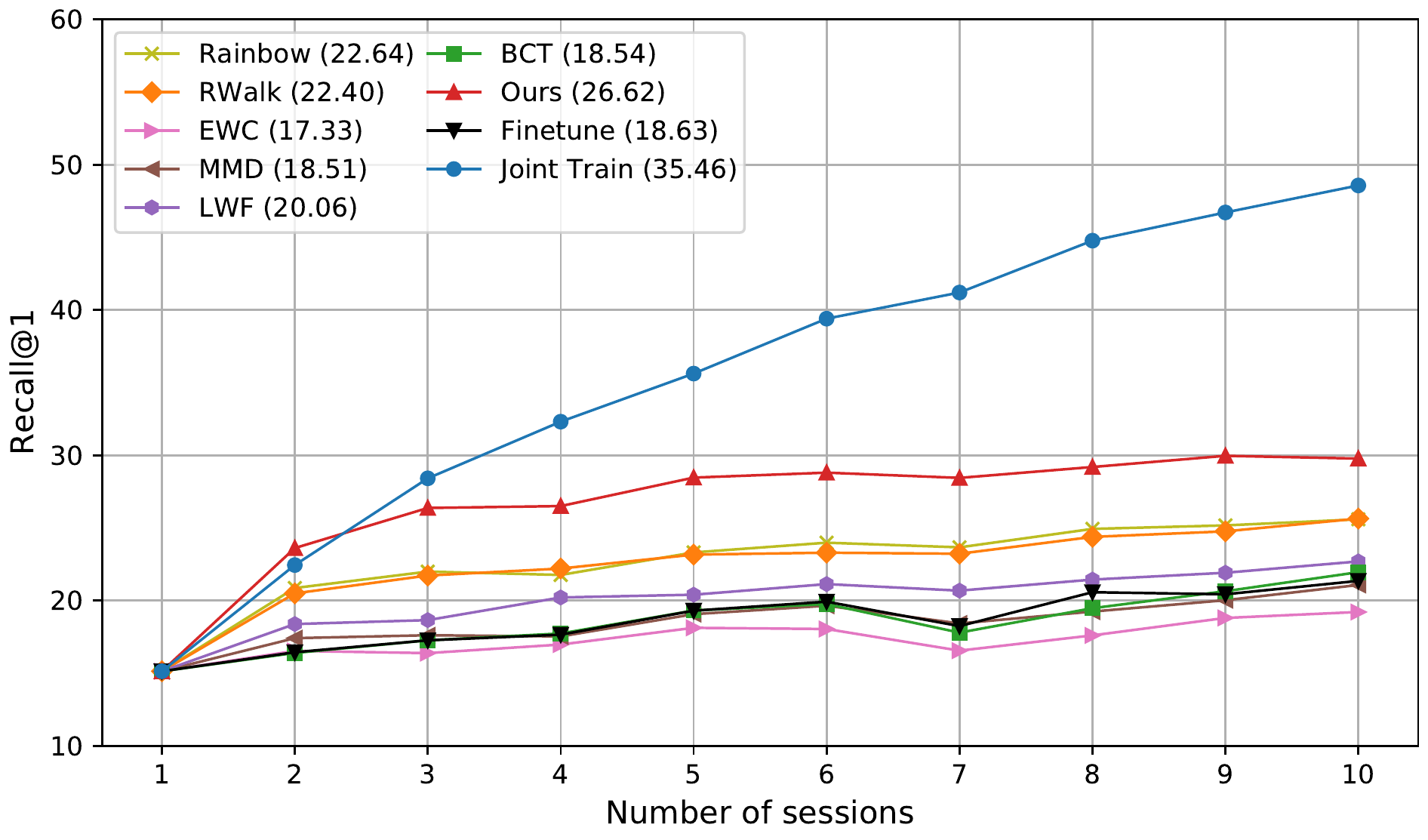}
                    \caption{Blurry Setup}
            \end{subfigure}
            \begin{subfigure}[t]{0.3\textwidth}
                    \includegraphics[width=\textwidth]{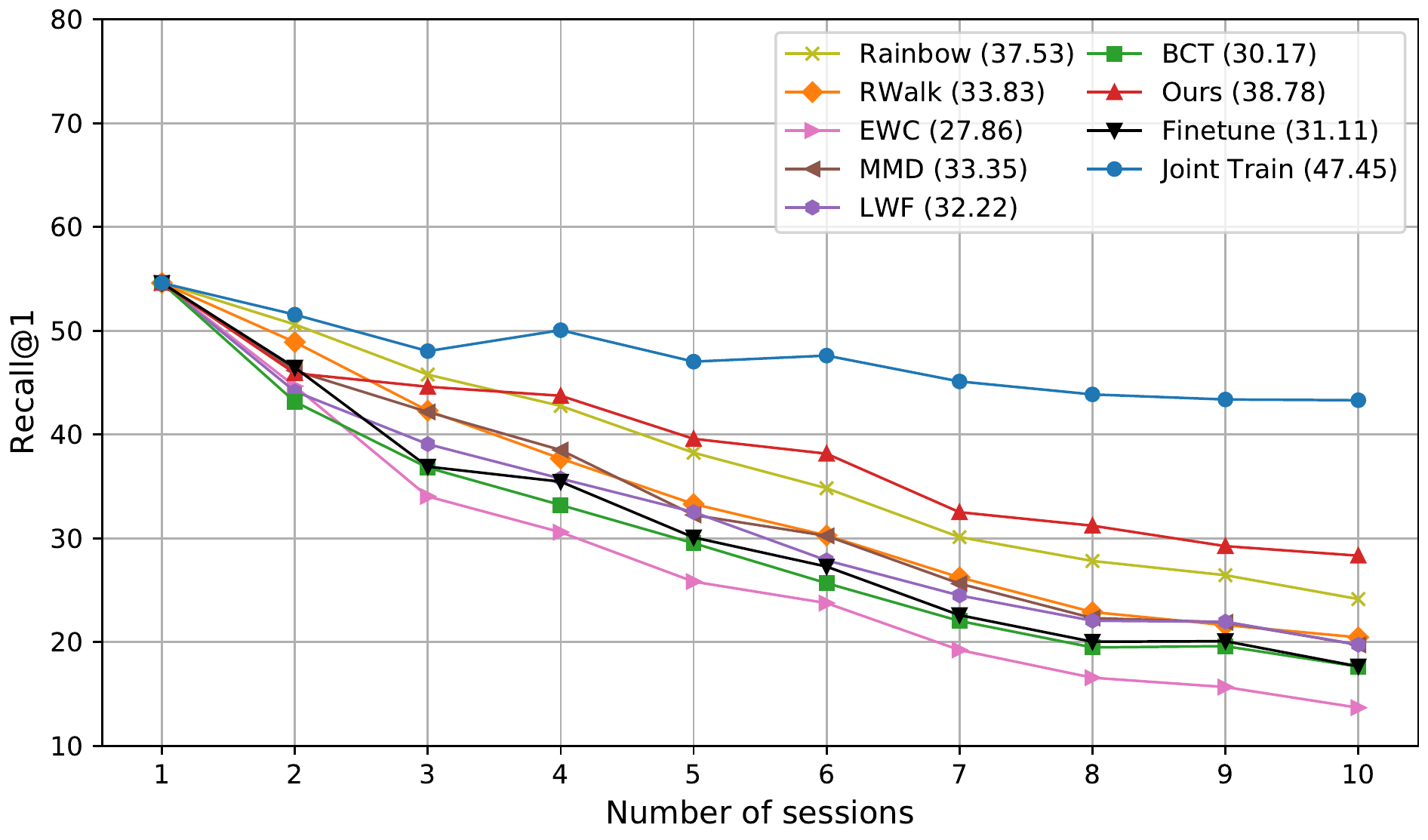}
                    \caption{General Setup}
            \end{subfigure}
    \vspace{-6pt}
    \caption{Recall@1 on Tiny ImageNet for the three setups, where Average Recall@1 across sessions is reported in parentheses. }
    \label{fig:tinyimagenetfigures}
\end{figure*}

%%%%%%%%%%%%%%%%%%%%%% 4.3 Results on the Coarse-grained Datasets
\subsection{Results on the Coarse-grained Datasets}

\noindent\textbf{Results on CIFAR100}:
We conduct a 5-session experiment on CIFAR100.
The results are shown in Fig.~\ref{fig:cifar100figures} (recall@1) and Table~\ref{tbl:averagerecall}(a) (\textbf{AR@K}).
For all three setups, except for \emph{Joint Train} that is an upper bound of this experiment, our CVS performs the best on both evaluation measures.
In the blurry and general-incremental setups, we observe that BCT performs even worse than \emph{Finetune}, a lower bound in this work.
Even though both old and new classes are shown for each incremental session, BCT would still require all samples seen previously for training to obtain satisfied results. 
On the other hand, we find that EWC attains almost the same performance as the lower bound. 
We attribute the bad performance to the uncertainty of importance weight estimation.
The runner-up is RWalk, which improves EWC by imposing constraints on the parameter space while avoiding forgetting via a replay-based mechanism. 
Such a mixed strategy works for classification but is not sufficiently well for backward-compatible retrieval. Instead, our CVS utilizes the inter- and neighbor-session information in extra, showing the efficacy under all setups.

\noindent\textbf{Results on Tiny ImageNet}:
We use Tiny ImageNet for simulating the 10-session CL scenario in this experiment.
To our knowledge, we're the first to perform such a long session sequence for backward-compatible retrieval in CL.
According to Fig.~\ref{fig:tinyimagenetfigures} and Table~\ref{tbl:averagerecall}(b), our method consistently outperforms the existing competitors on the three setups for this long sequence setting on retrieval.
The overall runner-up in this dataset becomes Rainbow, a data-replay method for classification.
Our CVS employs the replayed embedding to summarize a class across sessions for retrieval in addition to the data, and performs more favorably for all cases on recall@$1$ and \textbf{AR@K} particularly when $k$ is small.
For a large $k=4$ in the general setup, the performance is tie.
We attribute the gap is reduced since only one of the four retrieved labels has to be correct.

\noindent\textbf{Comparisons to Other Embedding Distillations}:
We conduct a comparison of our CVS to the techniques of metric learning from teacher~\cite{Yu_2019_CVPR,Park_2019_CVPR,Chen2018DarkRankAD} and knowledge distillation~\cite{Hinton_2015_NIPSW} by replacing $\mathbf{L^m}_{j;j+1}$ in Eq.~\ref{eqn:loss_combination} with different losses for neighbor-session model coherence based on CIFAR-100.
For a fair comparison, we carefully tune hyperparameter $\alpha$ at $\{1, 10\}$ for the better results at AR@1. 
Our 2-sample-3-embedding solution demonstrates the best position among competitors for all setups.
Compared to the runner-up results, we obtain CVS (71.47) v.s.~Anglewise-RKD~\cite{Park_2019_CVPR} (70.45) in disjoint setup, CVS(47.47) v.s.~Absolute MLKD~\cite{Yu_2019_CVPR} (46.9) in blurry setup, and CVS(73.95) v.s.~Dark Knowledge~\cite{Hinton_2015_NIPSW}(73.18) in general incremental setup. 
We detail the results in the supplementary material.

\noindent\textbf{Comparisons on Classification}:
As our CVS can produce the classification results via the NSoftmax layer too, we further compare the results with the CL classifier Rainbow~\cite{Bang_2021_CVPR} based on CIFAR-100 after finishing all sessions.
We have already followed the setting of~\cite{Bang_2021_CVPR} on both the class distribution ratios and replayed buffer size in the blurry setup.
The classification accuracy presented in~\cite{Bang_2021_CVPR} is $41.35\%$ with an online learning protocol (\ie, only a single epoch is allowed in learning).
We rerun the learner to converge and get the accuracy of $50.2\%$.
Surprisingly, CVS can achieve the $54.04\%$ classification accuracy that is even higher.
We owe the promising classification results to the replayed embedding that can serve as useful exemplars (like the principle of cross-batch learning~\cite{YangACMMM2016,Wang_2020_CVPR}) to further guide the training in our CVS. 
We conduct extra experiments on the other two setups (disjoint, general-incremental) 
and one additional CL classifier RWalk and obtain the results: CVS (50.62) v.s.~Rainbow (46.69) v.s.~RWalk (46.85) in disjoint setup, CVS (54.04) v.s.~Rainbow (50.2) v.s.~RWalk(50.89) in blurry setup, and CVS (55.49) v.s.~Rainbow (52.26) v.s.~RWalk(51.92) in general incremental setup.
The results demonstrate the efficacy of CVS for CL.

%%%%%%%%%%%%%%%%%%%%% RESULTS ON THE FINE-GRAINED DATASETS
\subsection{Results on the Fine-grained Datasets}

As for the fine-grained benchmarks, we consider the general-incremental setup only due to its practical usefulness.
We initialize the model with ImageNet pretrained weights as it is a common practice in the fine-grained retrieval benchmark.
We assume the first session presents samples from half of the classes from the dataset. 
It is a practical setting for a modern visual search system because a robust service should be well-trained to a certain extent before going online.
Due to the fine-grained limits on the data amount, we partition the Stanford Dog dataset into four, iNat-M into five, and Product-M into three sessions, respectively. 
The results are shown in Table~\ref{tbl:averagerecall}(c). 
Our CVS consistently outperforms the other approaches for all the fine-grained benchmarks and measures, except for AR@$k$ for $k=4$ on Stanford Dog Dataset. 
In sum, the evaluation results with other state-of-the-art approaches demonstrate that our approach is effective for both coarse- and fine-grained datasets on the general-incremental setup because we not only consider backward compatible embedding of the neighboring sessions but also long-term consistency with all the data of past sessions via replay embedding and data.
In addition, this backward compatible feature also saves the CL from expensive computational costs of gallery feature re-extraction as model updates during each session.

%%%%%%%%%%%%%%%%% DISCUSSION
\noindent\textbf{Discussion}:
A \emph{limitation} we find is that almost all methods (including ours) stagnate as the session expands, especially in the blurry setup. 
The margin between ours and upper bound enlarges gradually even though our method surpasses the remaining approaches. 
Therefore, there is still room for improvement in the long-term challenge.

%%%%%%%%%%%%%%%%%% ABLATION STUDY
\noindent\textbf{Ablation Study}:
Our CVS consists of three loss terms, $\mathbf{L}^{\mathbf{c}}$, $\mathbf{L}^{\mathbf{m}}$, and $\mathbf{L^d}$. 
To verify their effectiveness, we conduct the ablation study as shown in Table~\ref{tbl:ablationstudyaveragerecall}.
First, we find that  $\mathbf{L}^{\mathbf{c}}+\mathbf{L^d}$ has the greatest impact on the overall performance. 
It brings the gain in the range of 1.67\% to 11.37\% in AR@1 compared with $\mathbf{L}^{\mathbf{c}}$ alone. 
Second, the effect of $\mathbf{L}^{\mathbf{c}}+\mathbf{L}^{\mathbf{m}}$ is weaker despite a slight increase on CIFAR100 and Product-M compared to the one of $\mathbf{L}^{\mathbf{c}}$. 
Our finding suggests that adopting classification loss with distillation loss alone is insufficient. Therefore, seeking unification of additional information like $\mathbf{L^d}$ is practical in our setting.
To examine the performance gain from the consistency loss in more depth, we detach the exemplar replay technique from our method.  
We denote $\mathbf{L}^{\mathbf{c}}+\mathbf{L^d}$ \textbf{w/o replay} as the aforementioned case. 
The replay-based trick improves the plain version of the consistency loss on all datasets by a margin, especially for CIFAR100. Without reviewing exemplar data, the performance drop ranges between 1.08\% and 7.66\% in AR@1. Hence, integrating this design is essential for overall performance.

\begin{table}
\begin{center}
\resizebox{\columnwidth}{!}{
\begin{tabular}{|l|r|r|r|r|r|r|r|r|r|}
\hline
\multirow{2}{*}{} & \multicolumn{3}{c|}{CIFAR100}                                                     & \multicolumn{3}{c|}{Dog}                                                          & \multicolumn{3}{c|}{Product-M}                                                    \\ \cline{2-10} 
                  & \multicolumn{1}{l|}{AR@1} & \multicolumn{1}{l|}{AR@2} & \multicolumn{1}{l|}{AR@4} & \multicolumn{1}{l|}{AR@1} & \multicolumn{1}{l|}{AR@2} & \multicolumn{1}{l|}{AR@4} & \multicolumn{1}{l|}{AR@1} & \multicolumn{1}{l|}{AR@2} & \multicolumn{1}{l|}{AR@4} \\ \hline
$\mathbf{L}^{\mathbf{c}}$                & 60.79                     & 64.73                     & 68.14                     & 82.7                      & 88.32                     & 92.23                     & 70.99                     & 76.26                     & 81.16                     \\ \hline \hline
$\mathbf{L}^{\mathbf{c}}+\mathbf{L}^{\mathbf{m}}$               & 63.79                     & 68.13                     & 72.02                     & 82.65                     & 88.43                     & 92.29                     & 71.77                     & 76.92                     & 81.78                     \\ \hline
$\mathbf{L}^{\mathbf{c}}+\mathbf{L}^{\mathbf{d}} \: w/o \: replay$               & 64.5                      & 68.12                     & 71.14                     & 83.29                     & 88.77                     & 92.38                     & 73.96                     & 78.69                     & 83.02                     \\ \hline
$\mathbf{L}^{\mathbf{c}}+\mathbf{L}^{\mathbf{d}}$              & 72.16                     & 74.67                     & 76.7                      & 84.37                     & 89.07                     & 92.46                     & 75.6                      & 80.27                     & 84.33                     \\ \hline \hline
$\mathbf{L}^{\mathbf{c}}+\mathbf{L}^{\mathbf{m}}+\mathbf{L}^{\mathbf{d}}$             & 73.95                     & 76.73                     & 78.74                     & 84.71                     & 89.4                      & 92.61                     & 75.47                     & 80.36                     & 84.68                     \\ \hline
\end{tabular}
}
\vspace{-12pt}
\end{center}
\caption{Ablation study on each component, where $\mathbf{L}^{\mathbf{c}}$, $\mathbf{L}^{\mathbf{m}}$, and $\mathbf{L^d}$ are the losses of intra-session discrimination, neighbor-session model coherence, and inter-session data coherence, respectively. }
\label{tbl:ablationstudyaveragerecall}
\end{table}

%%%%%%%%%% CONCLUSION
\section{Conclusion}
\label{sec:conclusion}
In this work, we present a novel general incremental setup which allows the  new  gallery  set  of both seen and unseen classes incrementally added to the database and is closer to the real-world retrieval setup than the widely adopted disjoint and the recent blurry setups. 
Besides, we also propose a CL method for long-term visual search with backward consistent feature embedding. 
Our method acts as an extension of cross-batch memory to the cross-session memory for feature embedding learning in CL.
We introduce a 2-sample-3-embedding strategy in a triplet for distillation learning across neighbor sessions to enforce the model coherence.
The extensive experiments show that our approach achieves the state of the art performance in two coarse-grained classification and three fine-grained datasets under different incremental data distributions.

%%%%%%%%% REFERENCES
{\small
\bibliographystyle{ieee_fullname}
\bibliography{egbib}
}

\end{document}

%%%%%%%%% SUPPLEMENTARY - TITLE
\title{Supplementary Material \\ Continual Learning for Visual Search with Backward Consistent Feature Embedding}
\author{Timmy S. T. Wan\textsuperscript{1} \qquad Jun-Cheng Chen\textsuperscript{2} \qquad Tzer-Yi Wu\textsuperscript{3} \qquad Chu-Song Chen\textsuperscript{1,\thanks{\;indicates corresponding author.}}\\
National Taiwan University\textsuperscript{1} \qquad Academia Sinica\textsuperscript{2} \qquad ucfunnel Co. Ltd.\textsuperscript{3}\\
{\tt\small \{r08944004,chusong\}@csie.ntu.edu.tw, pullpull@citi.sinica.edu.tw, kenny.wu@ucfunnel.com}
}
\maketitle
\appendix
\renewcommand\thefigure{\thesection.\arabic{figure}}
\renewcommand\thetable{\thesection.\arabic{table}}
%%%%%%%%%%%%%%%%%%%%%%%%%% Contribution Review
\section{Contribution Review}
\begin{figure}[t]
   \includegraphics[width=1.05\linewidth]{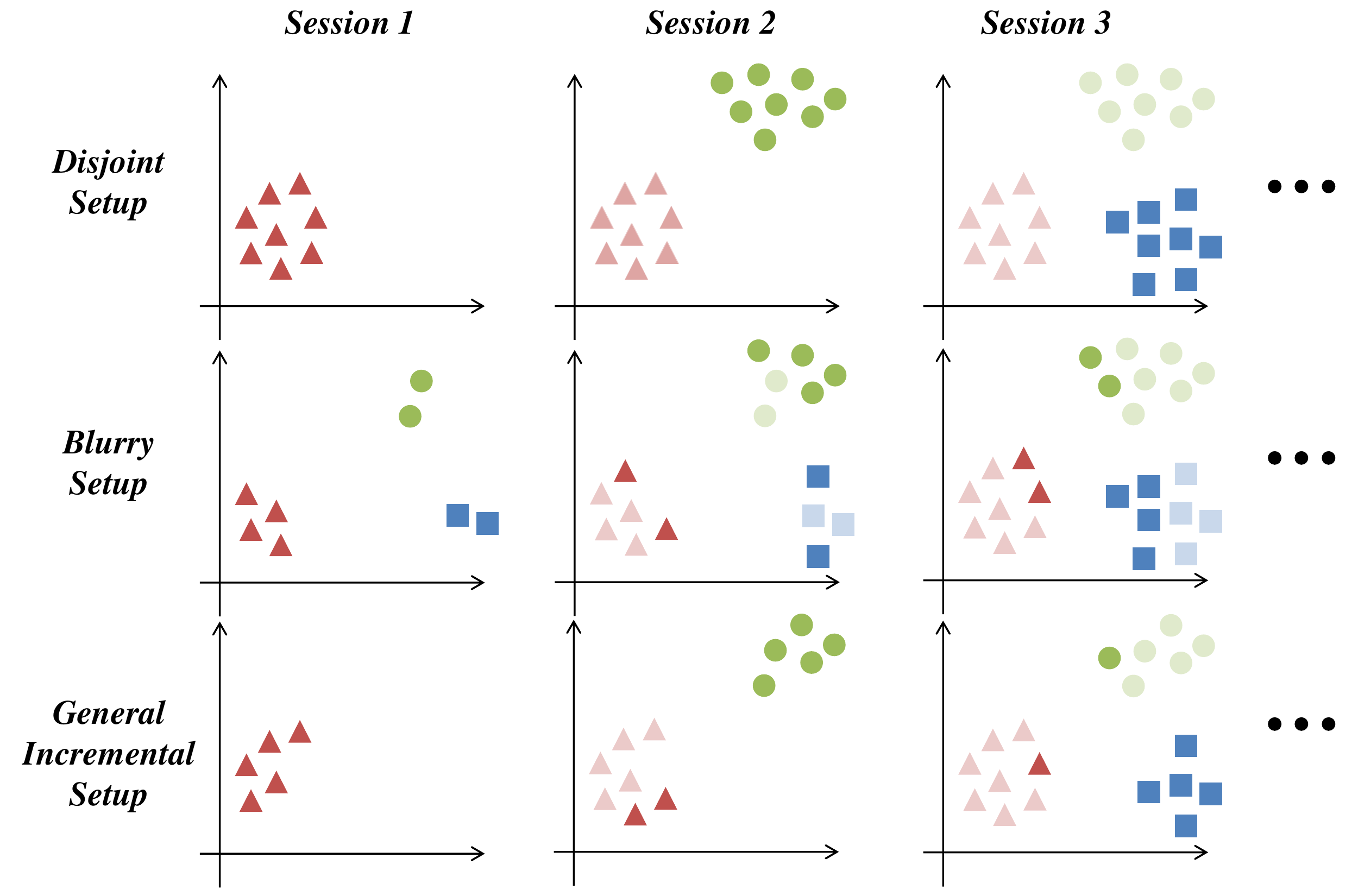}
   \caption{The widely adopted Disjoint setup (upper row) assumes the image categories mutually disjoint among sessions. The middle row shows the recent Blurry setup, where different sessions allow overlapping classes but all the classes are given initially; every session has a specific data distribution over the known classes. The bottom row shows our General Incremental setup, where the classes in a new session can be either old or novel.}
   \label{fig:suppvarioussetup}
 \end{figure}

\noindent\textbf{General Incremental Setup:}
Unlike previous works, we investigate a general case for CL (Fig.~\ref{fig:suppvarioussetup}).
Our setup considers the incremental classes of the disjoint setup and also covers overlapped classes of the blurry setup.
In a broad sense, it offers a more common situation for the CL research community on both classification and retrieval.

\noindent\textbf{Backward Consistent Feature Space Learning:}
We propose a novel continual learner for visual search allowing acquiring knowledge for unseen classes and making both the previous and the current feature space comparable without backfilling (\ie, re-extraction) of the previously processed gallery images.
We bridge this gap in three loss terms:
\begin{itemize}
  \item An inter-session data coherence loss learns from the history of all sessions by taking the extensible replayed embedding as a free supervision signal for guidance.
  \item A neighbor-session model coherence loss preserves the distance metric for the seen classes in both new and old sessions; it leverages a revised triplet loss with a new sampling strategy for distillation.
  \item An intra-session discrimination loss grasps knowledge from the novel categories using pointwise metric learning without loss of flexibility.
\end{itemize}

\begin{figure*}
\centering
  \includegraphics[width=0.9\linewidth]{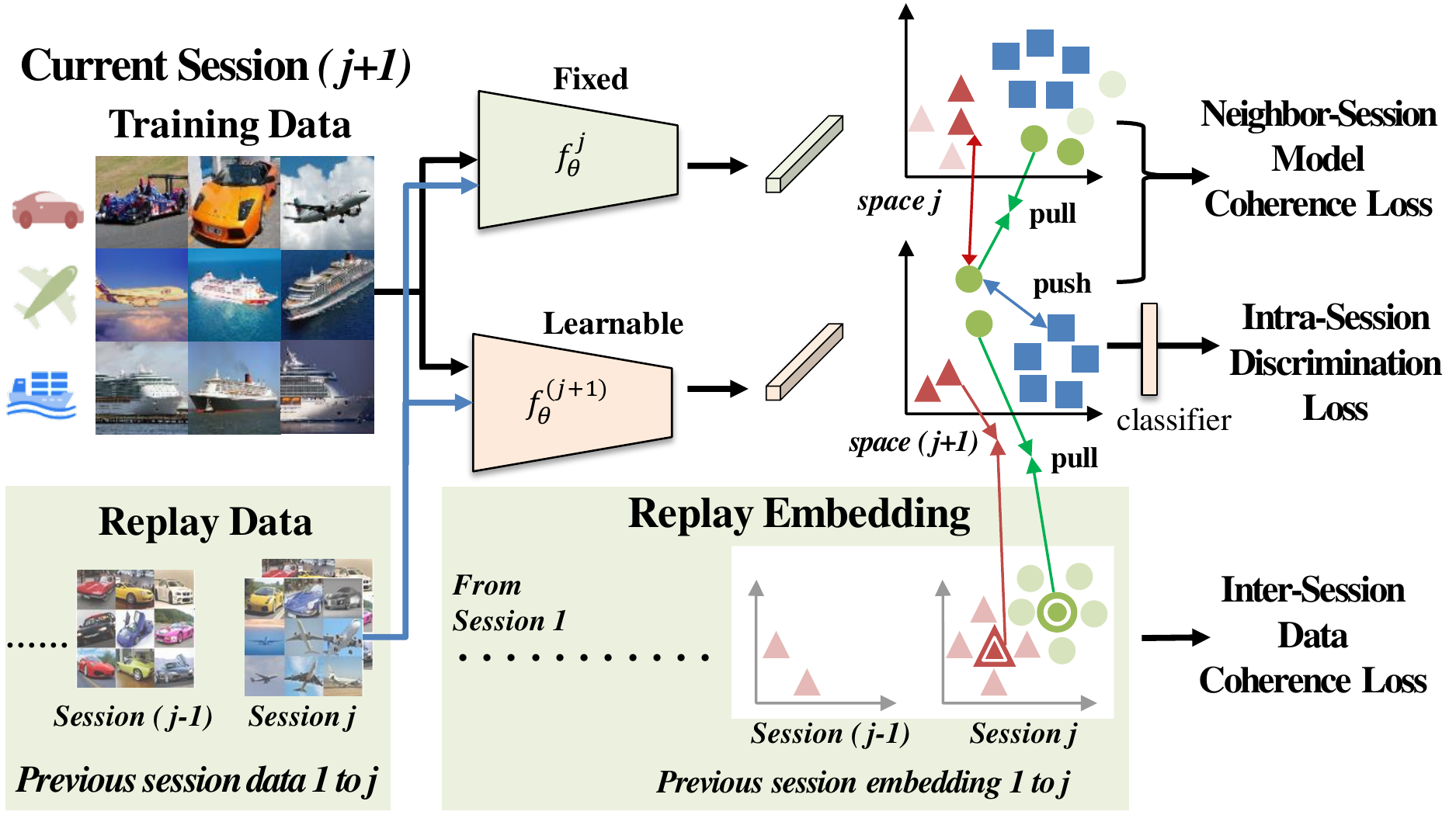}
  \caption{
  Overview of our CVS method for CL in General-Incremental setup with long-term backward embedding consistency. 
  In the current session $j+1$, the training data of the session (together with the replayed data under a budget control) are used in the three loss terms. 
  In addition, the replayed embeddings summarized from previous sessions $1:j$ serve as historically concentrated attractors to guide the inter-session data-coherence training; it acts as an extension of cross-batch memory~\cite{YangACMMM2016,Wang_2020_CVPR} to cross-session memory in CL.
  We introduce a 2-sample-3-embedding strategy in a triplet for distillation learning across neighbor sessions to enforce the model coherence.
  Note that we omit the replayed data to simplify the illustration.
  We use a L2-normalized embedding in classification~\cite{Zhai2019ClassificationIA} to provide the intra-session discriminating capability, and normalized embedding is adopted in all three loss terms.
  Our approach is simple but effective in all three CL setups, and we provide the first study on general-incremental setup in CL.
  }
  \label{fig:redraw_continual_senario}
\end{figure*}

An enlarged Fig.~2 of the main paper is provided for a more precise illustration, as shown in Fig.~\ref{fig:redraw_continual_senario}.
In the following, we complement more implementation details in Section~\ref{sec:Imp_details} and ablation studies in Section~\ref{sec:further_ablation}.

% FIGURE DIFFERENT SETUP

% FIGURE OUR METHOD

%%%%%%%%% visualization of fine-grained images
\begin{figure*}
	\centering
  \includegraphics[width=0.9\linewidth]{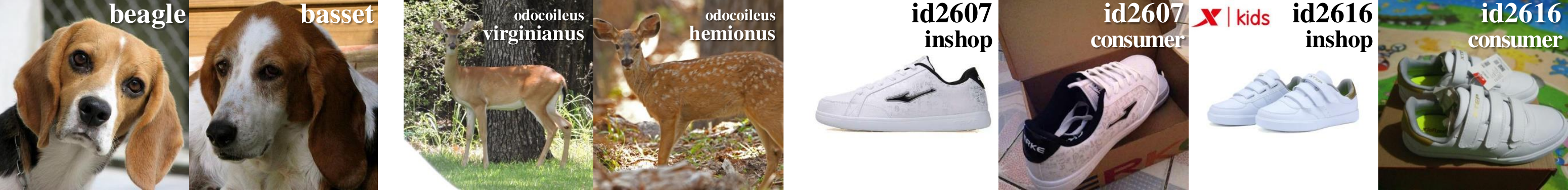}
	\caption{Sample images of fine-grained datasets. The first pair is from Stanford Dog, the latter pair is from iNaturalist 2017, and the final four images are from Product-10K.}
	\label{fig:datasetplot}
\end{figure*}

\begin{figure*}[htbp]
	\begin{center}
		\includegraphics[width=1.7\columnwidth]{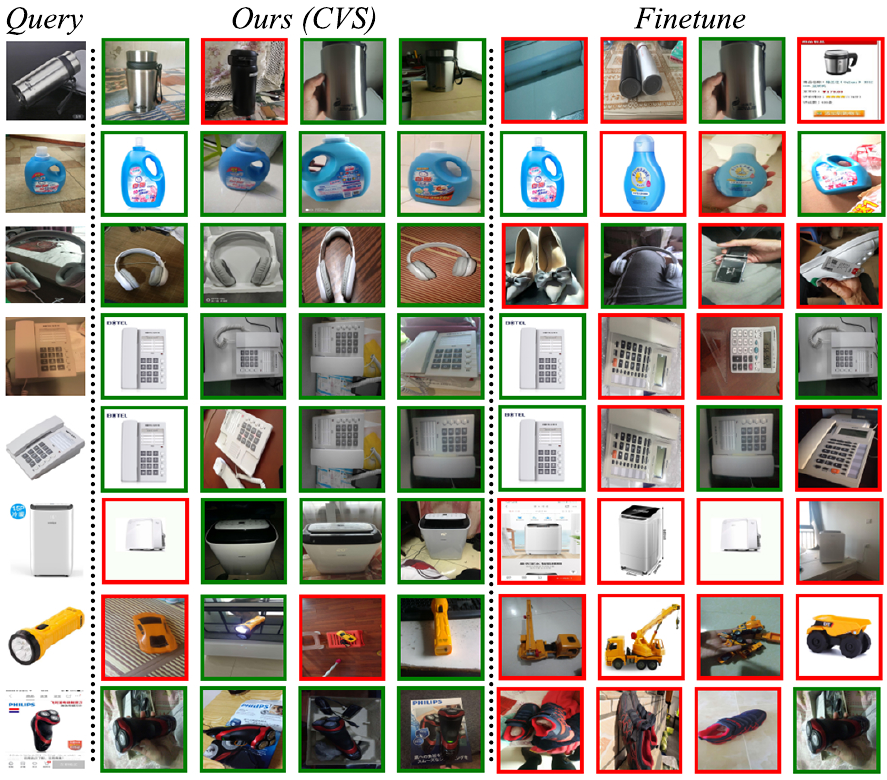}
		\caption{\textbf{Qualitative comparison} of the top-4 results using our method (CVS) and Finetune on the Product-M dataset.
		The correct and incorrect matches are highlighted in green and red, respectively.}
		\label{fig:retrievedresultproductm}
	\end{center}
	\vspace{-6pt}
\end{figure*}

%%%%%%%%%%%%%%%%%%% HYPERPARAMETER DETAILS
\section{Hyperparameter Details} \label{sec:Imp_details}
As presented in Section 4.1 of the main paper, we show the hyperparameter details as follows.
We implement all models with Pytorch~\cite{paszke2017automatic} using NVIDIA V100 GPUs.
To control the embedding size, we insert the fully connected layer with dimension $128$ before the final softmax layer given the network architecture.
The full experiments on the general incremental setup are shown in Tables~\ref{tbl:fullgeneral} and \ref{tbl:fullablation}.

\subsection{Details on the Coarse-grained Datasets}
The hyperparameters almost follow those in Rainbow~\cite{Bang_2021_CVPR} but with different batch sizes for Tiny ImageNet.
We train ResNet-18~\cite{He_2016_CVPR} over $256$ epochs with the batch size of $16$ and $64$ for CIFAR100 and Tiny ImageNet, respectively.
The networks are optimized using SGD with an initial learning rate of $0.03$, a momentum of $0.9$, and a weight decay of $0.0001$.
We adjust the learning rate in the range between $0.03$ and $0.0003$ by the cosine learning rate scheduler~\cite{LoshchilovH17}.
About data augmentation, training images from CIFAR100 are padded by $4$ pixels on all borders and then preprocessed through randomly cropping at $32\times32$, randomly horizontal flipping followed by AutoAug~\cite{Cubuk_2019_CVPR}.
For Tiny ImageNet, we follow the similar augmentation process but use randomly cropping at $64\times64$ and RandAug~\cite{Cubuk_2020_CVPR_Workshops} instead.

\subsection{Details on the Fine-grained Datasets}
Some fine-grained samples (Stanford Dog, iNat-M, and Product-M datasets) are shown in Fig.~\ref{fig:datasetplot}.
They have only subtle changes between classes and are more demanding for retrieval.
Following similar experimental settings mentioned in \cite{pmlr-v119-roth20a,Musgrave2020AML}, we finetune the ImageNet-pretrained ResNet-50 for $100$ epochs using SGD with a small fixed learning rate of $0.0001$.
The batch size is 64 by default but 32 for Product-M.
We follow \cite{pmlr-v119-roth20a} to preprocess the training images by randomly resizing and cropping them to $224\times224$ with random horizontal flipping.
At testing time, we emphasize the object by central cropping of $224\times224$ from the $256\times256$ resized image for feature extraction.

\subsection{Reimplementation details}
We re-implement the LWF~\cite{Li2018LearningWF}, MMD~\cite{ChenWei2020Oteo}, and BCT~\cite{Shen_2020_CVPR} in our experiments.
All loss terms are equal weighting to meet the balance between previously learned information and new knowledge.
For MMD, the maximum mean discrepancy loss is solely used for blurry setup, and an additional knowledge distillation loss is applied to the novel class data according to the original definition under the disjoint and general-incremental setups.
For BCT, we make some modifications to suit for different setups as it is not a CL solution and requires all old class samples collected so far.
We use all the old samples from the seen classes for the blurry and general-incremental setups, and employ the replayed data mined by iCaRL~\cite{Rebuffi_2017_CVPR} for the disjoint setup.

\subsection{Qualitative Study}
We demonstrate some qualitative results in Fig.~\ref{fig:retrievedresultproductm}.
Our method maintains the backward consistency and captures fine-grained characteristics of the particular object.
\Eg, the first row shows that our method can retrieve visually similar bottles, but Fine-tune yields the results with perturbation.

%%%%%%%%%%%%%%%%%%% Comparisons to Other Embedding Distillations

\section{Ablation study on different losses for neighbor-session model coherence}
As referenced in Section 4.3 of the main paper, we provide a complete experiment result as follows.

\noindent\textbf{Comparisons to Other Embedding Distillations:}
We perform an in-depth analysis of different metric losses for the loss term 
$\mathbf{L}^{\mathbf{m}}_{j;j+1}$, as shown in Table 3. 
Dark Knowledge~\cite{Hinton_2015_NIPSW} minimizes the KL divergence on the classifier side. 
Absolute MLKD~\cite{Yu_2019_CVPR} performs distillation at the penultimate layer output.
By estimating relational structural information given a mini-batch, RKD~\cite{Park_2019_CVPR} reduces the Huber loss using pairwise euclidean distance difference between two models (i.e., Distancewise RKD) and angle from three points (i.e., Anglewise RKD).
Following the same spirit, Relative MLKD~\cite{Yu_2019_CVPR} uses the difference of Frobenius norm instead, and DarkRank~\cite{Chen2018DarkRankAD} re-estimates the similarity ranks using listwise relationships.
Except for Dark Knowledge, we use their official Github implementation for fair comparisons.
We tune hyperparameter $\alpha$ at $\{1, 10\}$ for the best AR@1 value at the validation phase for importance weighting.
As can be seen in Table~\ref{tbl:avgrecalldistill}, our method provides the most favorable results at AR@$k$ when $k$ is small (1 or 2) on all setups, and only slightly inferior to Relative MLKD and DarkRank at AR@4 on the blurry and general-incremental setups, respectively.
\newpage

\noindent\textbf{Comparisons to Other Sample Mining Strategy:}
We examine different mining techniques because $\mathbf{L}^{\mathbf{m}}_{j;j+1}$ is computed based on the sampled triplets.
Instead of our easiest positive mining, we use the hardest positive mining (\ie, \textbf{BatchHard}, a hardest-positive-hardest-negative online mining strategy mentioned in ~\cite{Hermans2017InDO}) for fair comparisons (Table.~\ref{tbl:mining}).
We implement BatchHard with a balanced batch sampler to enforce each batch containing at least two samples per class for forming sufficient valid triplets.
Unfortunately, BatchHard obtains the worst result or even lower than the one with $\mathbf{L}^{\mathbf{m}}_{j;j+1}$ disabled.
We attribute this result to the misleading guidance due to a large variation between feature spaces; thus, mining triplets according to the cross-session distance is unreliable.
On the other hand, our strategy gains the best result by forming the positive samples using the outputs from two models without explicit mining.
Therefore, our triplet mining design is simple but effective, and easy to implement.

%%%%%%%%%%%% TABLE FULL RESULTS
\begin{table*}
\resizebox{\linewidth}{!}{
\begin{tabular}{|l|ccc|ccc|ccc|ccc|ccc|}
\hline
\multirow{2}{*}{} & \multicolumn{3}{c|}{CIFAR100}                                                              & \multicolumn{3}{c|}{Tiny ImageNet}                                                         & \multicolumn{3}{c|}{Dog}                                                                  & \multicolumn{3}{c|}{iNat-M}                                                                & \multicolumn{3}{c|}{Product-M}                                                             \\ \cline{2-16} 
                  & \multicolumn{1}{c|}{AR@1}           & \multicolumn{1}{c|}{AR@2}           & AR@4           & \multicolumn{1}{c|}{AR@1}           & \multicolumn{1}{c|}{AR@2}           & AR@4           & \multicolumn{1}{c|}{AR@1}           & \multicolumn{1}{c|}{AR@2}          & AR@4           & \multicolumn{1}{c|}{AR@1}           & \multicolumn{1}{c|}{AR@2}           & AR@4           & \multicolumn{1}{c|}{AR@1}           & \multicolumn{1}{c|}{AR@2}           & AR@4           \\ \hline
Joint Train       & \multicolumn{1}{c|}{81.97}          & \multicolumn{1}{c|}{84.6}           & 86.82          & \multicolumn{1}{c|}{47.45}          & \multicolumn{1}{c|}{51.94}          & 55.99          & \multicolumn{1}{c|}{86.98}          & \multicolumn{1}{c|}{91.08}         & 93.99          & \multicolumn{1}{c|}{75.85}          & \multicolumn{1}{c|}{79.97}          & 83.65          & \multicolumn{1}{c|}{79.36}          & \multicolumn{1}{c|}{83.83}          & 87.73          \\ \hline
Finetune          & \multicolumn{1}{c|}{60.79}          & \multicolumn{1}{c|}{64.73}          & 68.14          & \multicolumn{1}{c|}{31.11}          & \multicolumn{1}{c|}{35.87}          & 40.8           & \multicolumn{1}{c|}{82.7}           & \multicolumn{1}{c|}{88.32}         & 92.23          & \multicolumn{1}{c|}{67.75}          & \multicolumn{1}{c|}{72.68}          & 77             & \multicolumn{1}{c|}{70.99}          & \multicolumn{1}{c|}{76.26}          & 81.16          \\ \hline \hline
BCT               & \multicolumn{1}{c|}{58.31}          & \multicolumn{1}{c|}{62.2}           & 65.66          & \multicolumn{1}{c|}{30.17}          & \multicolumn{1}{c|}{34.64}          & 39.29          & \multicolumn{1}{c|}{81.73}          & \multicolumn{1}{c|}{87.49}         & 91.55          & \multicolumn{1}{c|}{67.34}          & \multicolumn{1}{c|}{72.07}          & 75.99          & \multicolumn{1}{c|}{70.48}          & \multicolumn{1}{c|}{75.67}          & 80.47          \\ \hline
LWF               & \multicolumn{1}{c|}{65.53}          & \multicolumn{1}{c|}{70.91}          & 75.31          & \multicolumn{1}{c|}{32.22}          & \multicolumn{1}{c|}{38.23}          & 44.56          & \multicolumn{1}{c|}{83.25}          & \multicolumn{1}{c|}{\textcolor{red}{\ul 89.29}}   & \textcolor{red}{\ul 93.04}    & \multicolumn{1}{c|}{68.51}          & \multicolumn{1}{c|}{73.71}          & 78.12          & \multicolumn{1}{c|}{72.95}          & \multicolumn{1}{c|}{78.38}          & 83.1           \\ \hline
MMD               & \multicolumn{1}{c|}{65.51}          & \multicolumn{1}{c|}{70.22}          & 74.33          & \multicolumn{1}{c|}{33.35}          & \multicolumn{1}{c|}{38.21}          & 43.06          & \multicolumn{1}{c|}{83.2}           & \multicolumn{1}{c|}{88.98}         & 92.71          & \multicolumn{1}{c|}{68.58}          & \multicolumn{1}{c|}{73.48}          & 77.79          & \multicolumn{1}{c|}{72.89}          & \multicolumn{1}{c|}{78.21}          & 82.96          \\ \hline
EWC               & \multicolumn{1}{c|}{60.89}          & \multicolumn{1}{c|}{64.86}          & 68.2           & \multicolumn{1}{c|}{27.86}          & \multicolumn{1}{c|}{32.67}          & 37.78          & \multicolumn{1}{c|}{81.64}          & \multicolumn{1}{c|}{88.7}          & 92.82          & \multicolumn{1}{c|}{66.3}           & \multicolumn{1}{c|}{71.4}           & 75.82          & \multicolumn{1}{c|}{66.01}          & \multicolumn{1}{c|}{72.53}          & 78.24          \\ \hline
RWalk             & \multicolumn{1}{c|}{\textcolor{red}{\ul 69.9}}     & \multicolumn{1}{c|}{\textcolor{red}{\ul 73.37}}    & \textcolor{red}{\ul 76.39}    & \multicolumn{1}{c|}{33.83}          & \multicolumn{1}{c|}{38.43}          & 42.97          & \multicolumn{1}{c|}{82.41}          & \multicolumn{1}{c|}{88.31}         & 92.43          & \multicolumn{1}{c|}{68.77}          & \multicolumn{1}{c|}{73.82}          & 78.16          & \multicolumn{1}{c|}{68.71}          & \multicolumn{1}{c|}{74.64}          & 80.13          \\ \hline
Rainbow           & \multicolumn{1}{c|}{68.4}           & \multicolumn{1}{c|}{71.56}          & 74.2           & \multicolumn{1}{c|}{\textcolor{red}{\ul 37.53}}    & \multicolumn{1}{c|}{\textcolor{red}{\ul 41.64}}    & \textcolor{red}{\ul 45.44}    & \multicolumn{1}{c|}{82.78}          & \multicolumn{1}{c|}{89.04}         & \textbf{93.23} & \multicolumn{1}{c|}{68.68}          & \multicolumn{1}{c|}{73.22}          & 77.21          & \multicolumn{1}{c|}{69.39}          & \multicolumn{1}{c|}{75.19}          & 80.34          \\ \hline \hline
Ours (CVS)        & \multicolumn{1}{c|}{\textbf{73.95}} & \multicolumn{1}{c|}{\textbf{76.73}} & \textbf{78.84} & \multicolumn{1}{c|}{\textbf{38.78}} & \multicolumn{1}{c|}{\textbf{42.38}} & \textbf{45.89} & \multicolumn{1}{c|}{\textbf{84.71}} & \multicolumn{1}{c|}{\textbf{89.4}} & 92.61          & \multicolumn{1}{c|}{\textbf{72.57}} & \multicolumn{1}{c|}{\textbf{76.39}} & \textbf{79.87} & \multicolumn{1}{c|}{\textbf{75.47}} & \multicolumn{1}{c|}{\textbf{80.36}} & \textbf{84.68} \\ \hline
CVS w/o replay    & \multicolumn{1}{c|}{67.61}          & \multicolumn{1}{c|}{71.3}           & 74.39          & \multicolumn{1}{c|}{36.39}          & \multicolumn{1}{c|}{40.32}          & 44.2           & \multicolumn{1}{c|}{\textcolor{red}{\ul 83.51}}    & \multicolumn{1}{c|}{88.96}         & 92.45          & \multicolumn{1}{c|}{\textcolor{red}{\ul 71.31}}    & \multicolumn{1}{c|}{\textcolor{red}{\ul 75.25}}    & \textcolor{red}{\ul 78.56}    & \multicolumn{1}{c|}{\textcolor{red}{\ul 74.19}}    & \multicolumn{1}{c|}{\textcolor{red}{\ul 78.95}}    & \textcolor{red}{\ul 83.27}    \\ \hline
\end{tabular}
}
\caption{Results on our general incremental setup. The champion is highlighted in bold and the runner-up is underlined in red.}
\label{tbl:fullgeneral}
\end{table*}

%%%%%%%%%%% TABLE FULL ABLATION
\begin{table*}
\resizebox{\linewidth}{!}{
\begin{tabular}{|l|ccc|ccc|ccc|ccc|ccc|}
\hline
                   & \multicolumn{3}{c|}{CIFAR100}                                                                                                                         & \multicolumn{3}{c|}{Tiny ImageNet}                                                                                                                   & \multicolumn{3}{c|}{Dog}                                                                                                                               & \multicolumn{3}{c|}{iNat-M}                                                                                                                               & \multicolumn{3}{c|}{Product-M}                                                                                                                           \\ \cline{2-16} 
\multirow{-2}{*}{} & \multicolumn{1}{c|}{AR@1}                               & \multicolumn{1}{c|}{AR@2}                               & AR@4                              & \multicolumn{1}{c|}{AR@1}                              & \multicolumn{1}{c|}{AR@2}                               & AR@4                              & \multicolumn{1}{c|}{AR@1}                               & \multicolumn{1}{c|}{AR@2}                               & AR@4                               & \multicolumn{1}{c|}{AR@1}                                  & \multicolumn{1}{c|}{AR@2}                               & AR@4                               & \multicolumn{1}{c|}{AR@1}                                 & \multicolumn{1}{c|}{AR@2}                               & AR@4                               \\ \hline
CVS: $\mathbf{L}^{\mathbf{c}}+\mathbf{L}^{\mathbf{m}}+\mathbf{L}^{\mathbf{d}}$        & \multicolumn{1}{c|}{\textbf{73.95}}                     & \multicolumn{1}{c|}{\textbf{76.73}}                     & \textbf{78.74}                    & \multicolumn{1}{c|}{\textbf{38.78}}                    & \multicolumn{1}{c|}{\textbf{42.38}}                     & \textbf{45.89}                    & \multicolumn{1}{c|}{\textbf{84.71}}                     & \multicolumn{1}{c|}{\textbf{89.4}}                      & \textbf{92.61}                     & \multicolumn{1}{c|}{\textcolor{red}{\ul 72.57}}    & \multicolumn{1}{c|}{\textbf{76.39}}                     & \textbf{79.87}                     & \multicolumn{1}{c|}{\textcolor{red}{\ul 75.47}}   & \multicolumn{1}{c|}{\textbf{80.36}}                     & \textbf{84.68}                     \\ \hline
CVS w/o replay     & \multicolumn{1}{c|}{67.61}                              & \multicolumn{1}{c|}{71.3}                               & 74.39                             & \multicolumn{1}{c|}{36.39}                             & \multicolumn{1}{c|}{40.32}                              & {\textcolor{red}{\ul 44.2}} & \multicolumn{1}{c|}{83.51}                              & \multicolumn{1}{c|}{88.96}                              & 92.45                              & \multicolumn{1}{c|}{71.31}                                 & \multicolumn{1}{c|}{75.25}                              & 78.56                              & \multicolumn{1}{c|}{74.19}                                & \multicolumn{1}{c|}{78.95}                              & 83.27                              \\ \hline
$\mathbf{L}^{\mathbf{c}}+\mathbf{L}^{\mathbf{d}}$                & \multicolumn{1}{c|}{\textcolor{red}{\ul 72.16}} & \multicolumn{1}{c|}{\textcolor{red}{\ul 74.67}} & {\textcolor{red}{\ul 76.7}} & \multicolumn{1}{c|}{\textcolor{red}{\ul 38.4}} & \multicolumn{1}{c|}{\textcolor{red}{\ul 41.22}} & 43.87                             & \multicolumn{1}{c|}{\textcolor{red}{\ul 84.37}} & \multicolumn{1}{c|}{\textcolor{red}{\ul 89.07}} & {\textcolor{red}{\ul 92.46}} & \multicolumn{1}{c|}{\textbf{72.61}} & \multicolumn{1}{c|}{\textcolor{red}{\ul 76.27}} & {\textcolor{red}{\ul 79.45}} & \multicolumn{1}{c|}{\textbf{75.6}} & \multicolumn{1}{c|}{\textcolor{red}{\ul 80.27}} & {\textcolor{red}{\ul 84.33}} \\ \hline
$\mathbf{L}^{\mathbf{c}}+\mathbf{L}^{\mathbf{d}}$ w/o replay     & \multicolumn{1}{c|}{64.5}                               & \multicolumn{1}{c|}{68.12}                              & 71.14                             & \multicolumn{1}{c|}{32.82}                             & \multicolumn{1}{c|}{36.2}                               & 39.55                             & \multicolumn{1}{c|}{83.29}                              & \multicolumn{1}{c|}{88.77}                              & 92.38                              & \multicolumn{1}{c|}{70.94}                                 & \multicolumn{1}{c|}{74.62}                              & 77.86                              & \multicolumn{1}{c|}{73.96}                                & \multicolumn{1}{c|}{78.69}                              & 83.02                              \\ \hline
$\mathbf{L}^{\mathbf{c}}+\mathbf{L}^{\mathbf{m}}$                & \multicolumn{1}{c|}{63.79}                              & \multicolumn{1}{c|}{68.13}                              & 72.02                             & \multicolumn{1}{c|}{32.1}                              & \multicolumn{1}{c|}{37.15}                              & 42.78                             & \multicolumn{1}{c|}{82.65}                              & \multicolumn{1}{c|}{88.43}                              & 92.29                              & \multicolumn{1}{c|}{67.79}                                 & \multicolumn{1}{c|}{72.64}                              & 76.97                              & \multicolumn{1}{c|}{71.77}                                & \multicolumn{1}{c|}{76.92}                              & 81.78                              \\ \hline
$\mathbf{L}^{\mathbf{c}}$                  & \multicolumn{1}{c|}{60.79}                              & \multicolumn{1}{c|}{64.73}                              & 68.14                             & \multicolumn{1}{c|}{31.11}                             & \multicolumn{1}{c|}{35.87}                              & 40.8                              & \multicolumn{1}{c|}{82.7}                               & \multicolumn{1}{c|}{88.32}                              & 92.23                              & \multicolumn{1}{c|}{67.75}                                 & \multicolumn{1}{c|}{72.68}                              & 77                                 & \multicolumn{1}{c|}{70.99}                                & \multicolumn{1}{c|}{76.26}                              & 81.16                              \\ \hline
\end{tabular}
}
\caption{Ablation results on our general incremental setup. The champion is highlighted in bold and the runner-up is underlined in red.}
\label{tbl:fullablation}
\end{table*}
\label{sec:further_ablation}

%%%%%%%%%%%% TABLE NEIGHBOR-SESSION MODEL COHERENCE
\begin{table*}[h!]
\begin{center}
\resizebox{0.7\linewidth}{!}{
\begin{tabular}{|l|r|r|r|r|r|r|r|r|r|}
\hline
\multirow{2}{*}{} & \multicolumn{3}{c|}{Disjoint}                                                  & \multicolumn{3}{c|}{Blurry}                                                    & \multicolumn{3}{c|}{General}                                                   \\ \cline{2-10} 
                  & \multicolumn{1}{l|}{AR@1} & \multicolumn{1}{l|}{AR@2} & \multicolumn{1}{l|}{AR@4} & \multicolumn{1}{l|}{AR@1} & \multicolumn{1}{l|}{AR@2} & \multicolumn{1}{l|}{AR@4} & \multicolumn{1}{l|}{AR@1} & \multicolumn{1}{l|}{AR@2} & \multicolumn{1}{l|}{AR@4} \\ \hline
None       & 69.85                    & 72.67                    & 74.58                    & 44.33                     & 46.14                    & 47.72                    & 72.16                    & 74.67                     & 76.7                    \\ \hline \hline
Dark Knowledge               & 69.41                    & 73.09                    & 76.05                    & 45.89                    & 47.76                    & 49.5                    & 73.18                    & 76.34                     & 78.68                    \\ \hline
Absolute MLKD               & 66.04                    & 70.36                    & 74.27                    & 46.9                    & 48.89                    & 50.77                    & 72.43                    & 75.72                    & 78.47                    \\ \hline
Relative MLKD               & 66.64                    & 71.2                    & 75.35                    & 44.1                    & 48.25                    & \textbf{52.33}                    & 70.91                    & 75.19                    & 78.92                    \\ \hline
Anglewise RKD               & 70.45                     & 72.65                    & 74.67                    & 45.07                    & 46.99                    & 48.67                    & 72.65                    & 74.67                    & 76.33                    \\ \hline
Distancewise RKD             & 70.1                    & 72.29                    & 74.11                    & 45.09                    & 47.01                    & 48.81                     & 72.83                     & 75.17                    & 77.14                    \\ \hline
Hard DarkRank           & 66.45                    & 71.03                    & 75.11                    & 44.8                    & 48.2                    & 51.35                    & 72.33                     & 76.15                    & \textbf{79.56}                     \\ \hline \hline
\textbf{Ours (CVS)}              & \textbf{71.47}                    & \textbf{74.8}                     & \textbf{77.51}                    & \textbf{47.47}                    & \textbf{49.86}                    & 52.17                    & \textbf{73.95}                    & \textbf{76.73}                    & 78.84                    \\ \hline
\end{tabular}
}
\vspace{-14pt}
\end{center}
\caption{Replace $\mathbf{L}^{\mathbf{m}}_{j;j+1}$ with different metric distillation losses on CIFAR100. \emph{None} disables $\mathbf{L}^{\mathbf{m}}_{j;j+1}$ for a simple baseline.}
\label{tbl:avgrecalldistill}
\end{table*}

%%%%%%%%%%% TABLE MINING STRATEGY
\begin{table*}[h!]
\begin{center}
\resizebox{0.65\linewidth}{!}{
\begin{tabular}{|l|r|r|r|r|r|r|r|r|r|}
\hline
\multirow{2}{*}{} & \multicolumn{3}{c|}{Disjoint}                                                  & \multicolumn{3}{c|}{Blurry}                                                    & \multicolumn{3}{c|}{General}                                                   \\ \cline{2-10} 
                  & \multicolumn{1}{l|}{AR@1} & \multicolumn{1}{l|}{AR@2} & \multicolumn{1}{l|}{AR@4} & \multicolumn{1}{l|}{AR@1} & \multicolumn{1}{l|}{AR@2} & \multicolumn{1}{l|}{AR@4} & \multicolumn{1}{l|}{AR@1} & \multicolumn{1}{l|}{AR@2} & \multicolumn{1}{l|}{AR@4} \\ \hline
None       & 69.85                    & 72.67                    & 74.58                    & 44.33                     & 46.14                    & 47.72                    & 72.16                    & 74.67                     & 76.7                    \\ \hline \hline
BatchHard           & 60.83                    & 63.91                    & 66.67                    & 39.24                    & 42.59                    & 45.7                    & 68.1                     & 71.02                    & 73.35                     \\ \hline \hline
\textbf{Ours (CVS)}              & \textbf{71.47}                    & \textbf{74.8}                     & \textbf{77.51}                    & \textbf{47.47}                    & \textbf{49.86}                    & \textbf{52.17}                    & \textbf{73.95}                    & \textbf{76.73}                    & \textbf{78.84}                    \\ \hline
\end{tabular}
}
\vspace{-14pt}
\end{center}
\caption{Compute $\mathbf{L}^{\mathbf{m}}_{j;j+1}$ with different mining strategies on CIFAR100.}
\label{tbl:mining}
\end{table*}

%%%%%%%%%%%%%%% TABLE MEMORY BUDGET
\begin{table*}[h!]
\begin{center}
\resizebox{0.55\linewidth}{!}{
\begin{tabular}{|c|ccc|ccc|}
\hline
                    & \multicolumn{3}{c|}{CIFAR100}                                   & \multicolumn{3}{c|}{Tiny ImageNet}                              \\ \hline
                    & \multicolumn{1}{c|}{AR@1}  & \multicolumn{1}{c|}{AR@2}  & AR@4  & \multicolumn{1}{c|}{AR@1}  & \multicolumn{1}{c|}{AR@2}  & AR@4  \\ \hline
RWalk (0.5\texttimes\;budget)   & \multicolumn{1}{c|}{66.9}  & \multicolumn{1}{c|}{70.32} & 73.59 & \multicolumn{1}{c|}{31.42} & \multicolumn{1}{c|}{36.39} & 41.41 \\ \hline
Rainbow (0.5\texttimes\;budget) & \multicolumn{1}{c|}{65.21} & \multicolumn{1}{c|}{69.33} & 72.85 & \multicolumn{1}{c|}{34.78} & \multicolumn{1}{c|}{39.8}  & 44.83 \\ \hline
\textbf{CVS (0.5\texttimes\;budget)}     & \multicolumn{1}{c|}{\textbf{71.99}} & \multicolumn{1}{c|}{\textbf{74.94}} & \textbf{77.25} & \multicolumn{1}{c|}{\textbf{37.72}} & \multicolumn{1}{c|}{\textbf{41.58}} & \textbf{45.59} \\ \hline \hline
RWalk (1\texttimes\;budget)   & \multicolumn{1}{c|}{69.9}  & \multicolumn{1}{c|}{73.37} & 76.39 & \multicolumn{1}{c|}{33.83} & \multicolumn{1}{c|}{38.43} & 42.97 \\ \hline
Rainbow (1\texttimes\;budget) & \multicolumn{1}{c|}{68.4}  & \multicolumn{1}{c|}{71.56} & 74.2  & \multicolumn{1}{c|}{37.53} & \multicolumn{1}{c|}{41.64} & 45.44 \\ \hline
\textbf{CVS (1\texttimes\;budget)}     & \multicolumn{1}{c|}{\textbf{73.95}} & \multicolumn{1}{c|}{\textbf{76.73}} & \textbf{78.84} & \multicolumn{1}{c|}{\textbf{38.78}} & \multicolumn{1}{c|}{\textbf{42.38}} & \textbf{45.89} \\ \hline 
\end{tabular}
}
\vspace{-14pt}
\end{center}
\caption{Results with different memory budgets on our general incremental setup. 1\texttimes\;budget represents the continual learner with a memory buffer size of $2000$ samples in CIFAR100 and $4000$ samples in TinyImageNet. 0.5\texttimes\;indicates using half the budget.}
\label{tbl:budgetsize}
\end{table*}

%%%%%%%%%%%% TABLE 3-run
\begin{table*}[h!]
\begin{center}
\resizebox{1.0\linewidth}{!}{
\begin{tabular}{|c|ccc|ccc|ccc|ccc|ccc|}
\hline
\multirow{2}{*}{} & \multicolumn{3}{c|}{CIFAR100}                                                  & \multicolumn{3}{c|}{Tiny ImageNet}                                             & \multicolumn{3}{c|}{Dog}                                                       & \multicolumn{3}{c|}{iNat-M}                                                    & \multicolumn{3}{c|}{Product-M}                                                 \\ \cline{2-16} 
                  & \multicolumn{1}{c|}{AR@1}       & \multicolumn{1}{c|}{AR@2}       & AR@4       & \multicolumn{1}{c|}{AR@1}       & \multicolumn{1}{c|}{AR@2}       & AR@4       & \multicolumn{1}{c|}{AR@1}       & \multicolumn{1}{c|}{AR@2}       & AR@4       & \multicolumn{1}{c|}{AR@1}       & \multicolumn{1}{c|}{AR@2}       & AR@4       & \multicolumn{1}{c|}{AR@1}       & \multicolumn{1}{c|}{AR@2}       & AR@4       \\ \hline
BCT               & \multicolumn{1}{c|}{58.01±0.26} & \multicolumn{1}{c|}{62.13±0.12} & 65.84±0.17 & \multicolumn{1}{c|}{30.11±0.43} & \multicolumn{1}{c|}{34.89±0.36} & 39.72±0.51 & \multicolumn{1}{c|}{81.77±0.07} & \multicolumn{1}{c|}{87.33±0.23} & 91.34±0.24 & \multicolumn{1}{c|}{67.17±0.31} & \multicolumn{1}{c|}{71.88±0.22} & 75.95±0.16 & \multicolumn{1}{c|}{70.55±0.1}  & \multicolumn{1}{c|}{75.83±0.17} & 80.67±0.4  \\ \hline
LWF               & \multicolumn{1}{c|}{65.39±0.24} & \multicolumn{1}{c|}{70.6±0.36}  & 75.06±0.29 & \multicolumn{1}{c|}{32.17±0.59} & \multicolumn{1}{c|}{38.3±0.41}  & 44.81±0.26 & \multicolumn{1}{c|}{83.45±0.33} & \multicolumn{1}{c|}{89.19±0.21} & 93.01±0.15 & \multicolumn{1}{c|}{68.34±0.3}  & \multicolumn{1}{c|}{73.71±0.04} & 78.2±0.07  & \multicolumn{1}{c|}{73.11±0.14} & \multicolumn{1}{c|}{78.42±0.04} & 83.01±0.19 \\ \hline
MMD               & \multicolumn{1}{c|}{65.15±0.32} & \multicolumn{1}{c|}{69.98±0.22} & 74.18±0.21 & \multicolumn{1}{c|}{33.3±0.3}   & \multicolumn{1}{c|}{38.47±0.27} & 43.57±0.46 & \multicolumn{1}{c|}{83.53±0.32} & \multicolumn{1}{c|}{89.01±0.15} & 92.65±0.06 & \multicolumn{1}{c|}{68.63±0.31} & \multicolumn{1}{c|}{73.6±0.26}  & 77.85±0.06 & \multicolumn{1}{c|}{72.81±0.27} & \multicolumn{1}{c|}{78.16±0.17} & 82.84±0.23 \\ \hline
EWC               & \multicolumn{1}{c|}{60.9±0.14}  & \multicolumn{1}{c|}{65.1±0.55}  & 68.76±0.92 & \multicolumn{1}{c|}{28.01±0.88} & \multicolumn{1}{c|}{33.21±0.66} & 38.55±0.66 & \multicolumn{1}{c|}{82.17±0.77} & \multicolumn{1}{c|}{89.1±0.68}  & \textbf{93.21±0.54} & \multicolumn{1}{c|}{66.18±0.1}  & \multicolumn{1}{c|}{71.53±0.12} & 76.02±0.17 & \multicolumn{1}{c|}{66.33±0.28} & \multicolumn{1}{c|}{72.53±0.04} & 78.04±0.17 \\ \hline
RWalk             & \multicolumn{1}{c|}{69.69±0.27} & \multicolumn{1}{c|}{73.16±0.19} & 76.23±0.14 & \multicolumn{1}{c|}{34.05±1.17} & \multicolumn{1}{c|}{38.77±1.07} & 43.48±0.98 & \multicolumn{1}{c|}{82.53±0.27} & \multicolumn{1}{c|}{88.67±0.33} & 92.81±0.34 & \multicolumn{1}{c|}{68.85±0.17} & \multicolumn{1}{c|}{73.9±0.11}  & 78.15±0.05 & \multicolumn{1}{c|}{69.05±0.31} & \multicolumn{1}{c|}{74.45±0.18} & 79.36±0.67 \\ \hline
Rainbow           & \multicolumn{1}{c|}{68.18±0.2}  & \multicolumn{1}{c|}{71.32±0.27} & 73.96±0.23 & \multicolumn{1}{c|}{37.47±0.51} & \multicolumn{1}{c|}{41.86±0.38} & 45.89±0.44 & \multicolumn{1}{c|}{83.06±0.29} & \multicolumn{1}{c|}{89.1±0.1}   & 93.15±0.09 & \multicolumn{1}{c|}{68.85±0.17} & \multicolumn{1}{c|}{73.43±0.33} & 77.41±0.36 & \multicolumn{1}{c|}{69.66±0.25} & \multicolumn{1}{c|}{74.91±0.3}  & 79.76±0.56 \\ \hline \hline
\textbf{Ours (CVS)}        & \multicolumn{1}{c|}{\textbf{73.81±0.15}} & \multicolumn{1}{c|}{\textbf{76.48±0.26}} & \textbf{78.58±0.27} & \multicolumn{1}{c|}{\textbf{38.56±0.33}} & \multicolumn{1}{c|}{\textbf{42.03±0.07}} & \textbf{46.01±0.23} & \multicolumn{1}{c|}{\textbf{84.64±0.07}} & \multicolumn{1}{c|}{\textbf{89.5±0.09}}  & 92.85±0.21 & \multicolumn{1}{c|}{\textbf{72.7±0.13}}  & \multicolumn{1}{c|}{\textbf{76.47±0.13}} & \textbf{79.74±0.23} & \multicolumn{1}{c|}{\textbf{75.77±0.35}} & \multicolumn{1}{c|}{\textbf{80.49±0.21}} & \textbf{84.7±0.19}  \\ \hline
\end{tabular}
}
\vspace{-14pt}
\end{center}
\caption{Average results of 3 runs on our general incremental setup.}
\label{tbl:multiplerun}
\end{table*}

%%%%%%%%%%%%%%%%%%%%%%%%%% More Technical Analysis
\section{More Technical Analysis}
\noindent\textbf{Results with different memory budgets:}
The experiment about the influence of the buffer size is provided in Table.~\ref{tbl:budgetsize}.
We observe that CVS consistently beats other replay-based methods on AR@1 using half the budget under the general-incremental setup.

\noindent\textbf{Multiple run results:}
We present the results of averaging 3 runs on all selected datasets.
With Table.~\ref{tbl:multiplerun}, most methods show no significant deviation ($<1\%$) except for RWalk in Tiny ImageNet.
But, this doesn't change any conclusions in our main paper.

\section*{Acknowledgement}
This work was supported in part by the MOST (Ministry of Science and Technology) under the grant MOST 110-2634-F-002-050, MOST 110-2634-F-006-022, and MOST 110-2221-E-002-185 -MY2.

%%%%%%%%% REFERENCES
{\small
\bibliographystyle{ieee_fullname}
\bibliography{egbib}
}